\documentclass[10pt,twocolumn,letterpaper]{article}

\usepackage{cvpr}              
\usepackage[accsupp]{axessibility} 

\usepackage[dvipsnames]{xcolor}

\usepackage{amssymb}
\usepackage{amsmath}
\usepackage{booktabs}
\usepackage{multirow}
\usepackage{tabularx}
\usepackage{color}
\usepackage{algorithm}
\usepackage{algorithmic}

\newcommand{\mbw}{\mathbf{w}}
\newcommand{\uds}{\underline}

\definecolor{cvprblue}{rgb}{0.21,0.49,0.74}
\usepackage[pagebackref,breaklinks,colorlinks,citecolor=cvprblue]{hyperref}

\def\paperID{12291} 
\def\confName{CVPR}
\def\confYear{2024}

\title{An Aggregation-Free Federated Learning for Tackling Data Heterogeneity}

\author{
Yuan Wang \quad Huazhu Fu \quad Renuga Kanagavelu \quad Qingsong Wei \quad Yong Liu \quad Rick Siow Mong Goh\\
Institute of High Performance Computing (IHPC)\\
Agency for Science, Technology and Research (A*STAR), Singapore
}

\begin{document}
\maketitle
\begin{abstract}
The performance of Federated Learning (FL) hinges on the effectiveness of utilizing knowledge from distributed datasets. Traditional FL methods adopt an aggregate-then-adapt framework, where clients update local models based on a global model aggregated by the server from the previous training round. This process can cause client drift, especially with significant cross-client data heterogeneity, impacting model performance and convergence of the FL algorithm. To address these challenges, we introduce FedAF, a novel aggregation-free FL algorithm. In this framework, clients collaboratively learn condensed data by leveraging peer knowledge, the server subsequently trains the global model using the condensed data and soft labels received from the clients. FedAF inherently avoids the issue of client drift, enhances the quality of condensed data amid notable data heterogeneity, and improves the global model performance. Extensive numerical studies on several popular benchmark datasets show FedAF surpasses various state-of-the-art FL algorithms in handling label-skew and feature-skew data heterogeneity, leading to superior global model accuracy and faster convergence.
\end{abstract}    
\section{Introduction}
\label{sec:intro}
\begin{figure}[t]
    \centering
    \begin{subfigure}{0.45\textwidth}
        \centering
        \includegraphics[width=0.9\textwidth]{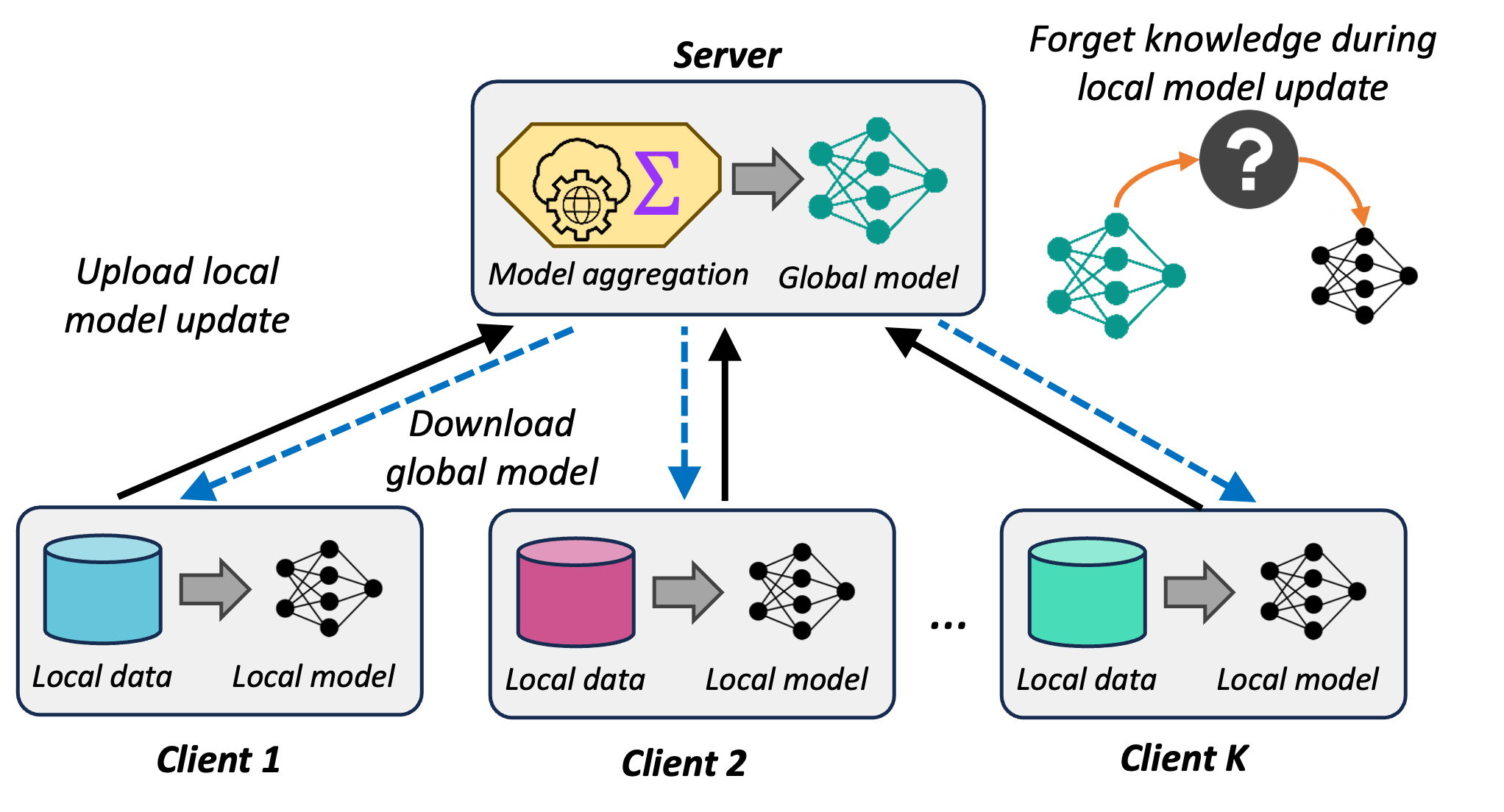}
        \caption{Aggregate-then-adapt FL approach}
        \label{fig:cover-a}
    \end{subfigure}
    \hfill
    \begin{subfigure}{0.45\textwidth}
        \centering
        \includegraphics[width=0.9\textwidth]{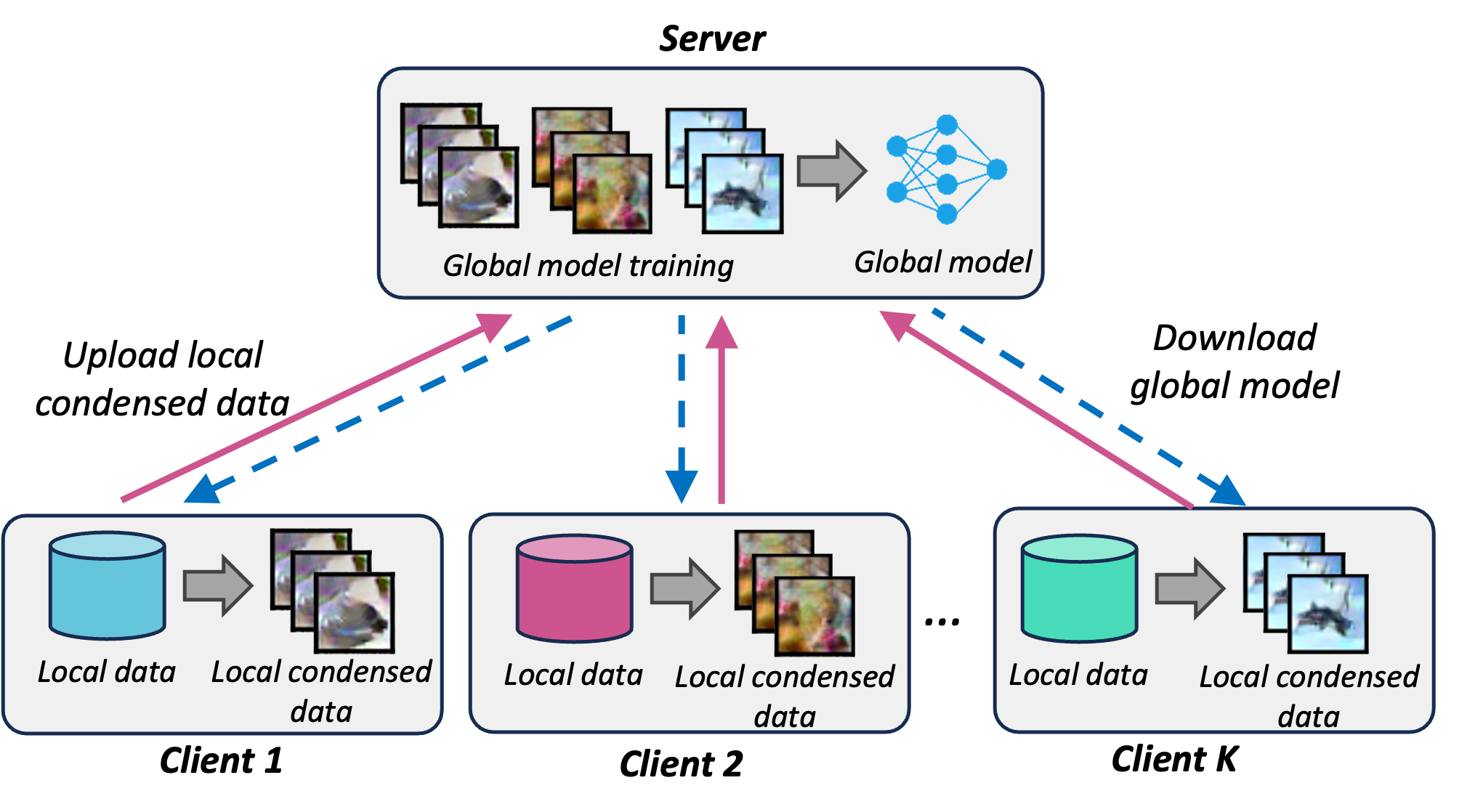}
        \caption{Aggregation-free FL approach}
        \label{fig:cover-b}
    \end{subfigure}
    \vskip -5 pt
    \caption{The conventional \textit{aggregate-then-adapt} approach (a) is prone to client drift in data-heterogeneous scenarios, as clients update a downloaded global model and risk forgetting prior knowledge. In contrast, the aggregation-free paradigm (b) has the server train the global model directly using condensed synthetic data learned and shared by clients, which circumvents the client drift issue.}
    \vskip -15 pt
    \label{fig:cover}
\end{figure}
Federated Learning (FL) algorithms typically follow an iterative \textit{aggregate-then-adapt} paradigm in which clients use their local private data to refine a global model provided by a central server. These locally updated models are subsequently returned to and aggregated by the server, where the global model is updated through averaging the local models according to predetermined rules \cite{mcmahan:17}. Given the decentralized nature of client data in FL, substantial variations in the clients' data distribution are common. This scenario results in non-Independent and Identically Distributed (non-IID) data across clients, often referred to as data heterogeneity. Such heterogeneity presents considerable challenges to the convergence of FL algorithms, primarily due to the significant drift in local learning paths among clients \cite{karimireddy:20, wang:21}. This phenomenon, known as client drift, can significantly decrease the accuracy of the global model. \cite{zhao:18, zhu:21, li:22}. 

Most existing efforts to address the challenge of data heterogeneity have focused either on modifying the local model training with additional regularization terms \cite{li:20, li:21moon, karimireddy:20, acar:21feddyn, gao22:feddc} or on utilizing alternative server-side aggregation or model update schemes \cite{hsu:19fedavgm, wang:20fednova, lin20:feddf, zhu:21fedgen, chen:21fedbe, zhang:22fine}. 
Nevertheless, these methods remain constrained by the conventional \textit{aggregate-then-adapt} framework, as depicted in Figure~\ref{fig:cover-a}. In scenarios with strong cross-client non-IID data distribution, the fine-tuning of the global model over local data becomes susceptible to catastrophic forgetting. This means that clients tends to forget the knowledge learned in the global model and diverge away from the stationary point of global learning objective when they update the model individually \cite{huang:22learn, feng:23visualprompt}. 

Recent works on data condensation \cite{wang:18dd, zhao:21gm, zhao21:dsa, cazenavette:22mtt, wang:22cafe, zhao:23dm} 
suggest an \textit{aggregation-free} paradigm that has the potential to overcome the limitations mentioned above in FL with privacy preservation.
As depicted in Figure~\ref{fig:cover-b}, in this new framework, each client first learns a compact set of synthetic data (i.e., the condensed data) for each class and then shares the learned condensed data with the server. The server then utilizes the received condensed data to directly update the global model. 
However, in the current research on the \textit{aggregation-free} method \cite{xiong:22feddm, liu:23fedmk}, two critical open challenges emerge: First, significant cross-client data heterogeneity can compromise the quality of locally condensed data, adversely affecting the global model training. Second, relying exclusively on condensed data for global model training can result in reduced convergence performance and robustness, particularly when the quality of the received condensed data is sub-optimal.

Motivated by the above research gap, this paper presents FedAF, a novel \textit{aggregation-free} FL algorithm tailored to combat data heterogeneity. At the heart of our research is the question of how to optimally harness the inherent knowledge in each client's original data to enhance both local data condensation and global model training. To achieve this, we first introduce a \textit{collaborative data condensation} scheme. In this scheme, clients condense their local dataset by minimizing a loss function that integrates a standard distribution matching loss \cite{zhao:23dm} with an additional regularization term based on Sliced Wasserstein Distance (SWD). This regularization aligns the local knowledge distribution with the broader distribution across other clients, granting individual clients a more comprehensive perspective for data condensation. Furthermore, to train the global model with superior performance, we incorporate a \textit{local-global knowledge matching} scheme. This approach enables the server to utilize not only the condensed data shared by clients but also soft labels extracted from their data, thereby refining and stabilizing the training process. As a result, the global model retains more knowledge from earlier rounds, leading to enhanced overall convergence performance.

Extensive experiments demonstrate that FedAF can consistently deliver superior model performance and accelerated convergence speed, outperforming several state-of-the-art FL algorithms across various degrees of data heterogeneity. For instance, on CIFAR10, we achieve up to 25.44\% improvement in accuracy and 80\% improvement in convergence speed, compared with the state-of-the-art methods. In summary, our contributions are threefold as follows:
\begin{itemize} 
    \item We propose a novel \textit{aggregation-free} FL algorithm, termed FedAF, to tackle the challenge of data heterogeneity. Unlike traditional approaches that aggregate local model gradients, FedAF updates the global model using client-condensed data, thereby effectively circumventing client drift issues.
    \item We introduce a collaborative data condensation scheme to enhance the quality of condensed data. By employing a Sliced Wasserstein Distance-based regularization, this scheme allows each client to leverage the broader knowledge in the data of other clients, a feature not adequately explored in existing literature.
    \item We further present a local-global knowledge matching scheme which equips the server with soft labels extracted from client data for enhanced global insights. This scheme supplements the condensed data received from clients, thereby facilitating improved model accuracy and accelerating convergence speed.
\end{itemize}

\section{Background and Related Works}
\label{sec:background_related_works}

\begin{figure*}[!t]
\centering
\includegraphics[width=1\textwidth]{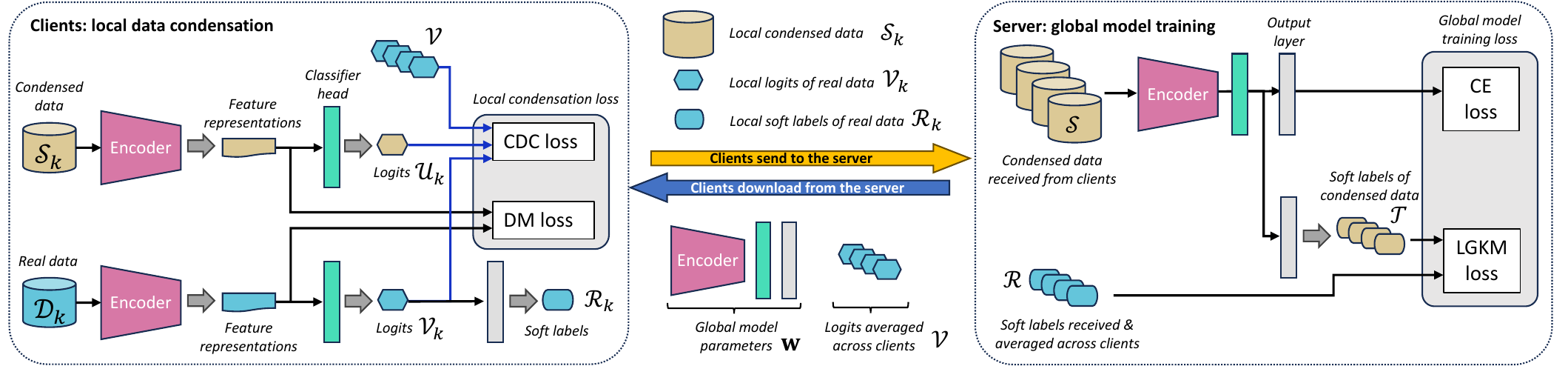}
\caption{Overview of FedAF's workflow. Left: Clients download the global model $\mathbf{w}$ and the class-wise mean logits $\mathcal{V}$, averaged from $\mathcal{V}_k$ at the server. They then update the condensed data $\mathcal{S}_k$ using a combination of Distribution Matching (DM) loss and Collaborative Data Condensation (CDC) loss, with local real data $\mathcal{D}_k$ and $\mathcal{V}$ as inputs. Right: The server updates the global model $\mathbf{w}$ by employing both cross-entropy loss and Local-Global Knowledge Matching (LGKM) loss. This utilizes both condensed data $\mathcal{S}_k$ and soft labels $\mathcal{R}_k$ received from each client $k \in \{1,2,\dots, N\}$. The entire process iterates over a pre-defined number of communication rounds.}
\vskip -15 pt
\label{fig:overallflow}
\end{figure*}

\noindent\textbf{FL Algorithms for Heterogeneous Data.} 
The foundational FedAvg algorithm \cite{mcmahan:17}, widely used in FL, calculates the global model by averaging the local models from each client. Among its variants, FedAvgM \cite{hsu:19fedavgm} adds server-side Nesterov momentum to enhance the global model update. FedNova \cite{wang:20fednova} normalizes aggregation weights based on the amount of local computation. FedBN \cite{li:21fedbn} specifically excludes batch normalization layer parameters from global aggregation. FedProx \cite{li:20fedprox} integrates a proximal term in local training loss to mitigate the issue of client drift, while SCAFFOLD \cite{karimireddy:20} employs variance reduction and a control variate technique to directly address client drift. FedDyn \cite{acar:21feddyn} allows clients to update the regularization in their local training loss to align more closely with the global empirical loss. In contrast, FedDC \cite{gao22:feddc} proposes learning a drift variable to actively mitigate discrepancies between local and global model parameters. MOON \cite{li:21moon} uses model-contrastive regularization to foster similarity in feature representations between the global and local models. Meanwhile, FedDF \cite{lin20:feddf} and FedBE \cite{chen:21fedbe} focus on knowledge distillation-based model fusion and Bayesian model ensemble, respectively, to transfer knowledge into the global model. To forego the need for an unlabelled transfer dataset, FedGen \cite{zhu:21fedgen} enables the server to learn and share a generator model with clients, facilitating knowledge distillation from the global model to local models through generated feature representations.

\medskip
\noindent\textbf{Data Condensation.} 
Recent years have witnessed the rise of data condensation (or data distillation) techniques. These methods aim to compress a large training dataset into a significantly smaller set of synthetic data, enabling models trained on this condensed dataset to achieve performance comparable to those trained on the original dataset. For example, \cite{wang:18dd} explores a bi-level learning framework to learn the condensed data, allowing models trained on it to minimize the loss on the original data. \cite{zhao:21gm} matches the gradients produced by training models on both the original and condensed data. This approach is further enhanced by \cite{zhao21:dsa}, which applies differential Siamese augmentation to enable the learning of more informative condensed data. Differing from single-step gradient matching, \cite{cazenavette:22mtt} suggests matching multiple steps of training trajectories resulting from both the original and condensed data. To reduce the complexity of the expensive bi-level optimization used in earlier works, \cite{zhao:23dm} introduces a distribution matching (DM) protocol. In this approach, condensed data is optimized to closely match the distribution of the original data in the latent feature space. Building upon DM, \cite{wang:22cafe} further develops this concept by aligning layer-wise features between the real and condensed data.

Unlike Generative Adversarial Networks (GANs) \cite{goodfellow:14gan}, which focus on generating realistic-looking images, the objective of data condensation methods is to boost data efficiency by creating highly informative synthetic training samples. A recent study by \cite{dong22:dcprivacy} indicates that condensed data not only provides robust visual privacy, but also ensures that models trained on this data exhibit resistance to membership inference attacks.

The potential of condensed data has recently drawn interest in the FL community, prompting efforts to combine data condensation with FL within an \text{aggregation-free} framework. FedDM \cite{xiong:22feddm} implements a DM-based data condensation on the client side, with the server using condensed data from clients to approximate the original global training loss in FL. Employing the condensation method in \cite{wang:18dd} as the backbone, \cite{liu:23fedmk} introduces a dynamic weighting strategy for local data condensation and enhances global model training with pseudo data samples obtained from a conditional generator. However, both of these works do not adequately investigate how to utilize knowledge from other clients to improve the quality of local condensed data and the performance of the global model. While \cite{liu:23fedmk} allows the sharing of condensed data among clients, it relies on an assumption that all clients possess data of the same class, which may not be true under strong cross-client data heterogeneity.
\section{Notations and Preliminaries}
\label{sec:preliminary}

To facilitate a clearer understanding of our proposed method, we begin by introducing some essential notations and preliminaries.

\medskip
\noindent\textbf{Federated Learning.}
FL is a decentralized machine learning framework where $K$ clients jointly learn a global model parameterized by $\mathbf{w}$ without uploading local raw data $\mathcal{D}=\{\mathcal{D}_1, \mathcal{D}_2, \dots, \mathcal{D}_K\}$, where $\mathcal{D}_k=\{(x_k^i, y_k^i)\}_{i=1}^{|\mathcal{D}_k|}$ is the local data owned by client $k$ and $|\cdot|$ refers to the number of samples in $\mathcal{D}_k$. These clients are then set to learn the model parameter $\mbw$ by solving the following problem collaboratively:
\begin{equation}
\small
    \arg\min_{\mbw}\mathcal{L}_\text{glob}(\mbw) = \sum_{i=k}^{K} p_k \mathcal{L}_k(\mbw, \mathcal{D}_k),
\label{eq:glob_loss}
\end{equation}
\vskip -5 pt
\noindent where $\mathcal{L}_k(\mbw, \mathcal{D}_k)$ is the local empirical loss of client $k$, which is given by:
\begin{equation}
\small
    \mathcal{L}_k(\mbw, \mathcal{D}_k) = \frac{1}{|\mathcal{D}_k|}\sum_{i=1}^{|\mathcal{D}_k|} \ell_k(\mbw, x^i_k, y^i_k),
\label{eq:loc_loss}
\end{equation}
\vskip -5 pt
\noindent where $\ell_k(\cdot)$ denotes the training loss such as cross-entropy loss. Weighting coefficient $p_k$ associated with client $k$ is proportional to $|\mathcal{D}_k|$ and normalized by $p_k = \frac{|\mathcal{D}_k|}{|\mathcal{D}|}$
so that $\sum_{k=1}^{K}p_k=1$. 

\medskip
\noindent\textbf{Data Condensation with Distribution Matching.}
Suppose a client $k$ is tasked with learning a set of local condensed data denoted as $\mathcal{S}_k$. The client first initializes each class of condensed data by sampling from the local original data $\mathcal{D}_k$ or Gaussian noise. Subsequently, the client employs a feature extractor, denoted as $h_{\mbw}(\cdot)$, which comprises all the layers preceding the last fully connected layer in a backbone model parameterized by $\mbw$, and extracts the feature representations from each class $c \in \{0, 1, \dots, C-1\}$ of data in $\mathcal{D}_k$ and $\mathcal{S}_k$. The means of these feature representations are computed as below,
\begin{equation}
\small
    \mu_{k,c}^\text{real} = \frac{1}{N_{k,c}} \sum_{j=1}^{N_{k,c}} h_{\mbw} (x_{k,c}^j), \mu_{k,c}^\text{syn} = \frac{1}{M_{k,c}} \sum_{j=1}^{M_{k,c}} h_{\mbw} (\Tilde{x}_{k,c}^j).
\end{equation}
\vskip -5 pt
Here $x_{k,c}^j \in \mathcal{D}_k$ and $\Tilde{x}_{k,c}^j \in \mathcal{S}_k$ represent the $j$-th sample of class $c$ drawn from $\mathcal{D}_k$ and $\mathcal{S}_k$, respectively. $N_{k,c}$ refers to the number of samples of class $c$ in $\mathcal{D}_k$, whereas $M_{k,c}$ denotes the number of samples of class $c$ in $\mathcal{S}_k$. Then client $k \in \{0, 1, \dots, K-1\}$ can learn and update $\mathcal{S}_k$ by optimizing the DM loss as follows,
\begin{equation}
\small
    \arg \min_{\mathcal{S}_k} \mathcal{L}_\text{DM} (\mathcal{S}_k, \mathcal{D}_k) = \sum_{c=0}^{C-1} \| \mu_{k,c}^\text{real} (\mathcal{D}_k) - \mu_{k,c}^\text{syn} (\mathcal{S}_k)\|^2.
\label{eq:dm_loss}
\end{equation}

\section{The Proposed Method}
\label{sec:method}
The overall mechanism of our proposed FedAF is illustrated in Figure~\ref{fig:overallflow}. In each learning round, clients first engage in collaborative data condensation, updating their local condensed data. They then share this condensed data and soft labels with the server. Subsequently, the server uses this information to update the global model.

\medskip
\noindent\textbf{Collaborative Data Condensation.}
Other than the extraction of feature representations needed by DM loss in \eqref{eq:dm_loss}, we allow clients to compute class-wise mean logits and related soft labels from their local original data $\mathcal{D}_k$. With the latest global model updated by the server as the backbone network, clients perform the following:
\begin{equation}
\small
    \mathbf{v}_{k,c}  = \frac{1}{N_{k,c}}\sum_{j=1}^{N_{k,c}} f_{\mbw}(x_{k,c}^j),
\label{eq:loc_mean_logits}
\end{equation}
\vskip -5 pt
\noindent where $\mathbf{v}_{k,c}$ refers to the mean logit of class $c$ computed by client $k$, function $f_{\mbw}(\cdot)$ denotes the global model parameterized by $\mbw$ without the last softmax layer. Similarly, clients can compute class-wise mean logits of local condensed data $\mathbf{u}_{k,c}$ by
\begin{equation}
\small
    \mathbf{u}_{k,c} = \frac{1}{M_{k,c}}\sum_{j=1}^{M_{k,c}} f_{\mbw}(\Tilde{x}_{k,c}^j).
\label{eq:loc_mean_logits_syn}
\end{equation}
\vskip -5 pt
To ease the presentation, we adopt the following nations for every client $k$:
\begin{equation}
\small
\begin{split}
    \mathcal{V}_k &= [\mathbf{v}_{k,0}, \mathbf{v}_{k,1}, \dots, \mathbf{v}_{k,C-1}], \\
    \mathcal{U}_k &= [\mathbf{u}_{k,0}, \mathbf{u}_{k,1}, \dots, \mathbf{u}_{k,C-1}].
\end{split}
\end{equation}
Upon having the class-wise mean logits obtained, clients share $\mathcal{V}_k$ with the server. The server then updates the global class-wise mean logits $\mathbf{v}_{c}$ by
\begin{equation}
\small
    \mathbf{v}_c \leftarrow \frac{1}{K}\sum_{k=0}^{K-1} \mathbf{v}_{k,c}.
\label{eq:glob_mean_logits}
\end{equation}

We then empower clients to learn and update the local condensed data $\mathcal{S}_k$ by downloading the following from the server
\begin{equation}
    \mathcal{V} = [\mathbf{v}_{0}, \mathbf{v}_{1}, \dots, \mathbf{v}_{C-1}],
    \label{eq:collective_mean_logits}
\end{equation}
and minimizing a local training loss as below:
\begin{equation}
\small
\begin{split}
    \mathcal{L}_\text{loc} & \bigl(\mathcal{S}_k, \{\mathcal{D}_k\}_{k=0}^{K-1} \bigr)  = \mathcal{L}_\text{DM} (\mathcal{S}_k, \mathcal{D}_k) \\
     &+\lambda_\text{loc} \sum_{c=0}^{C-1} \mathcal{F}\Bigl(\mathbf{u}_{k,c}(\mathcal{S}_k),\mathbf{v}_c(\mathcal{D}_1, \mathcal{D}_2, \dots, \mathcal{D}_K)\Bigr),   
\end{split}
\label{eq:cdc_loss}
\end{equation}
where we refer to the second term on the right-hand-side of \eqref{eq:cdc_loss} as the collaborative data condensation (CDC) loss. 
Without loss of generality, we let $\mathcal{F}(\cdot, \cdot)$ in the above loss function be a distance metric to promote the alignment between the local knowledge $\mathbf{u}_{k,c}$ and the global knowledge $\mathbf{v}_c$. Inspired by recent studies \cite{arjovsky:17wgan, kolouri:18sliced, deshpande:18swdgan}, we select the Sliced Wasserstein Distance (SWD) as the choice of $\mathcal{F}(\cdot, \cdot)$. SWD serves as an effective approximation of the exact Wasserstein distance \cite{bonneel:15swd, kolouri:16swd}, enabling the efficient capture of discrepancies between the knowledge distributions of locally condensed data and the original data owned by other clients. 
This approach allows clients not only to match the distributions of condensed and original data in the latent feature space but also ensures a matching across clients in logit space. By leveraging the CDC loss as a regularization term, each client can learn condensed data with the assistance of global insights shared by their peer clients, thereby avoiding the pitfall of biased matching towards their local data and facilitating the learning of higher-quality condensed data. 

\medskip
\noindent\textbf{Local-Global Knowledge Matching.}
The data condensation process inevitably leads to a certain degree of information loss from the original data. Consequently, relying solely on the condensed data received from clients for global model training might result in limited convergence performance or even instability. To address this, we introduce a local-global knowledge matching approach, enabling the server to harness a broader spectrum of knowledge about the original data distributed across clients.

To be more specific, we let clients compute local class-wise soft labels about its original data, represented by 
\begin{equation}
\small
    \mathcal{R}_k = [\mathbf{r}_{k,0}, \mathbf{r}_{k,1}, \dots, \mathbf{r}_{k,C-1}],
\label{eq:loc_soft_labels}
\end{equation}
\vskip -5 pt
\noindent where $\mathbf{r}_{k,c}=\sigma_{\mbw}(\mathbf{v}_{k,c}, \tau)$ and $\sigma_{\mbw}(\cdot,\tau)$ denotes the last softmax layer of the global model $f_{\mbw}$ softened by a temperature $\tau$. Clients then share $\mathcal{R}_k$ with the server in addition to the local condensed data $\mathcal{S}_k$, the server subsequently averages the shared local soft labels for each class as below
\begin{equation}
\small
    \mathbf{r}_c \leftarrow \frac{1}{K}\sum_{k=0}^{K-1} \mathbf{r}_{k,c}.
\label{eq:avg_class_knowledge}
\end{equation}
\vskip -5 pt
At the same time, the server uses the received condensed data $\mathcal{S}=\{\mathcal{S}_1, \mathcal{S}_2, \dots, \mathcal{S}_K\}$ to compute global class-wise soft labels $\mathbf{t}_c$ through
\begin{equation}
\small
    \mathbf{t}_c = \sigma_{\mbw} \biggl(\frac{1}{|\mathcal{S}_c|}\sum_{j=1}^{|\mathcal{S}_c|}f_{\mbw}(\Tilde{x}^j_c), \tau \biggr),
\label{eq:glob_syn_soft_label}
\end{equation}
\vskip -5 pt
\noindent where $\Tilde{x}^j_c \in \mathcal{S}_c$ and $\mathcal{S}_c$ denotes the condensed data received by the server and belongs to class $c$. Then we empower the server to match $\mathbf{t}_c$ with $\mathbf{r}_c$ using a Kullback–Leibler (KL) divergence-based regularization term and propose the following loss function for training the global model:
\begin{equation}
\small
    \mathcal{L}_\text{glob}(\mbw, \mathcal{S}) = \mathcal{L}_\text{CE}(\mbw, \mathcal{S}) + \lambda_\text{glob} \mathcal{L}_\text{LGKM}(\mbw, \mathcal{S}),
\label{eq:glob_training_loss}
\end{equation}
where $\mathcal{L}_\text{CE}(\mbw, \mathcal{S})$ denotes the cross-entropy (CE) loss and $\mathcal{L}_\text{LGKM}(\mbw, \mathcal{S})$ refers to the local-global knowledge matching (LGKM) regularization, which is given by
\begin{equation}
\small
    \mathcal{L}_\text{LGKM}(\mbw, \mathcal{S}) = \frac{1}{2} \Bigl(D_\text{KL}(\mathcal{R}||\mathcal{T}) + D_\text{KL}(\mathcal{T}||\mathcal{R}) \Bigr).
\end{equation}
$\mathcal{R}$ and $\mathcal{T}$ in the above equation are defined as
\begin{equation}
\small
    \mathcal{R} = [\mathbf{r}_0, \mathbf{r}_1, \dots, \mathbf{r}_{C-1}], \mathcal{T} = [\mathbf{t}_0, \mathbf{t}_1, \dots, \mathbf{t}_{C-1}].    
\label{eq:loc_glob_soft_labels}
\end{equation}
\vskip -5 pt
Augmented by the local-global knowledge matching, the server gains a more comprehensive understanding of the original data distributed across clients. By ensuring the global model better retains knowledge acquired in prior rounds, this enhancement not only stabilizes the global model updates but also curbs potential overfitting, especially when dealing with less-optimal condensed data.

\medskip
\noindent\textbf{Model Re-sampling.}
We also employ a global model re-sampling approach in each step of the local data condensation.  Specifically, we interpolate the global model received from the server as below
\begin{equation}
\label{eq:model_resample}
\small
    \mbw \leftarrow \gamma \mbw + (1-\gamma) \Tilde{\mbw},
\end{equation}
\vskip -5 pt
\noindent where $\Tilde{\mbw}$ represents randomly sampled model parameters. This model re-sampling technique aims to mitigate overfitting when learning condensed data by interpolating between the parameters of the global model learned in the previous round and a benign random perturbation.
\section{Experiments}
\label{sec:results}

\noindent\textbf{Datasets.} We conduct experiments to evaluate and benchmark the performance of our proposed methods on both label-skew data heterogeneity and feature-skew data heterogeneity. For label-skew scenarios, we adopt FashionMNIST (FMNIST) \cite{xiao:17fmnist}, CIFAR10, and CIFAR100 \cite{krizhevsky:09cifar}. The FMNIST dataset consists of 70,000 grayscale images of fashion products that fall into 10 categories. The CIFAR-10 and CIFAR-100 datasets are both collections of 60,000 color images. CIFAR10 images are grouped into 10 classes with 6,000 images per class, while CIFAR100 categorizes image into 100 classes with 600 images per class. For feature-skew scenario, we use DomainNet \cite{peng2019moment} which contains 600,000 million images in six domains: Clipart, Infograph, Painting, Quickdraw, Real, and Sketch, each domain has data of 345 classes.

\medskip
\noindent\textbf{Baselines.} We compare FedAF against the \textit{aggregate-then-adapt} baselines, including typical FedAvg \cite{mcmahan:17} and various state-of-the-art FL algorithms designed for handling data heterogeneity: FedProx \cite{li:20fedprox}, FedBN \cite{li:21fedbn}, MOON \cite{li:21moon}, FedDyn \cite{acar:21feddyn}, and FedGen \cite{zhu:21fedgen}. We also consider the prior work that also employs \textit{aggregation-free} FL, namely FedDM \cite{xiong:22feddm}, to demonstrate the effectiveness of our proposed collaborative data condensation and local-global knowledge matching schemes.

\begin{table*}
    \centering
    \setlength{\tabcolsep}{3.5pt}
    \fontsize{9pt}{10pt}\selectfont
    \begin{tabular}{l|c|c|c|c|c|c|c|c|c}
        \toprule
        \multirow{2}{*}{Methods} & \multicolumn{3}{c|}{$\alpha=0.02$} & \multicolumn{3}{c|}{$\alpha=0.05$} & \multicolumn{3}{c}{$\alpha=0.1$} \\
        & FMNIST & CIFAR10 & CIFAR100 & FMNIST & CIFAR10 & CIFAR100 & FMNIST & CIFAR10 & CIFAR100 \\
        \midrule
        FedAvg  & 56.50$\pm$5.55 & 39.71$\pm$1.15 & 30.80$\pm$2.20 & 69.14$\pm$5.84 & 46.51$\pm$3.07 & 33.37$\pm$0.75 & 82.19$\pm$5.67  & 56.15$\pm$4.62 & 39.97$\pm$1.53 \\
        FedProx & 60.38$\pm$5.00 & 36.46$\pm$5.39 & 30.82$\pm$0.80 & 69.33$\pm$4.12 & 45.83$\pm$2.23 & 36.61$\pm$1.44 & 81.56$\pm$4.52  & 58.54$\pm$1.87 & 40.45$\pm$1.53 \\
        FedBN   & 58.26$\pm$4.28 & 36.53$\pm$2.52 & 29.73$\pm$1.73 & 72.91$\pm$4.69 & 45.13$\pm$2.18 & 33.73$\pm$2.15 & 77.33$\pm$3.07 & 57.67$\pm$3.21 & 39.84$\pm$0.20 \\
        MOON    & 51.33$\pm$7.00 & 33.32$\pm$1.13 & 33.41$\pm$0.70 & 71.41$\pm$4.08 & 47.41$\pm$4.59 & 37.90$\pm$0.80 & 81.61$\pm$2.68  & 57.62$\pm$4.99 & 40.24$\pm$0.68 \\
        FedDyn  & 69.79$\pm$5.04 & 45.73$\pm$3.98 & 35.01$\pm$2.07 & 75.19$\pm$5.49 & 57.68$\pm$1.84 & 39.10$\pm$0.34 & 84.73$\pm$2.74  & 59.97$\pm$2.20 & 41.81$\pm$1.46 \\
        FedGen  & 61.44$\pm$2.07 & 36.61$\pm$1.06 & 29.20$\pm$2.09 & 75.48$\pm$1.83 & 42.72$\pm$2.11 & 33.56$\pm$3.91 & 82.29$\pm$2.53  & 58.17$\pm$2.84 & 40.23$\pm$1.06 \\
        \midrule
        FedDM   & 85.36$\pm$0.96 & 60.28$\pm$0.82 & 44.15$\pm$0.30 & 86.08$\pm$0.68 & 62.97$\pm$0.96 & 46.27$\pm$0.98 & 86.65$\pm$0.31  & 64.88$\pm$0.35 & 47.05$\pm$0.13 \\
        FedAF   & \textbf{87.53}$\pm$\textbf{0.32} & \textbf{65.15}$\pm$\bf{0.86} & \textbf{48.71}$\pm$\textbf{0.33} & \textbf{87.29}$\pm$\textbf{0.23} & \textbf{67.50}$\pm$\bf{0.76} & \textbf{49.49}$\pm$\textbf{0.33} & \textbf{87.91}$\pm$\textbf{0.41} & \textbf{69.11}$\pm$\textbf{0.86} & \textbf{50.61}$\pm$\textbf{0.26} \\
        \bottomrule    
    \end{tabular}
    \vskip -5 pt
    \caption{Comparison of global model accuracy achieved with various FL algorithms on FMNIST, CIFAR10, and CIFAR100 datasets. FedAF consistently outperforms all baseline methods across three degrees of data heterogeneity.}
    \vskip -10 pt
    \label{tab:overview}
\end{table*}

\begin{figure*}
    \centering
    \begin{subfigure}{0.3\linewidth}
        \includegraphics[width=1.0\textwidth]{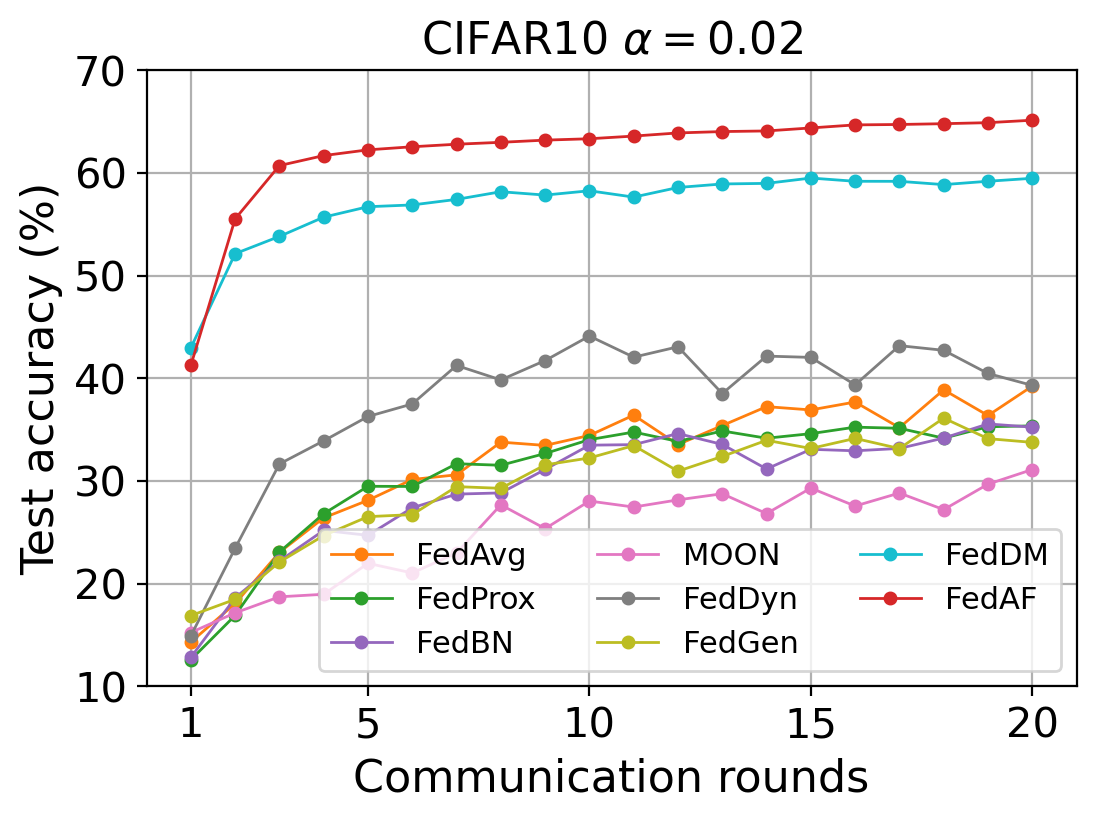}
        \caption{}
        \label{fig:result-a}
    \end{subfigure}
    \hfill
    \begin{subfigure}{0.3\linewidth}
        \includegraphics[width=1.0\textwidth]{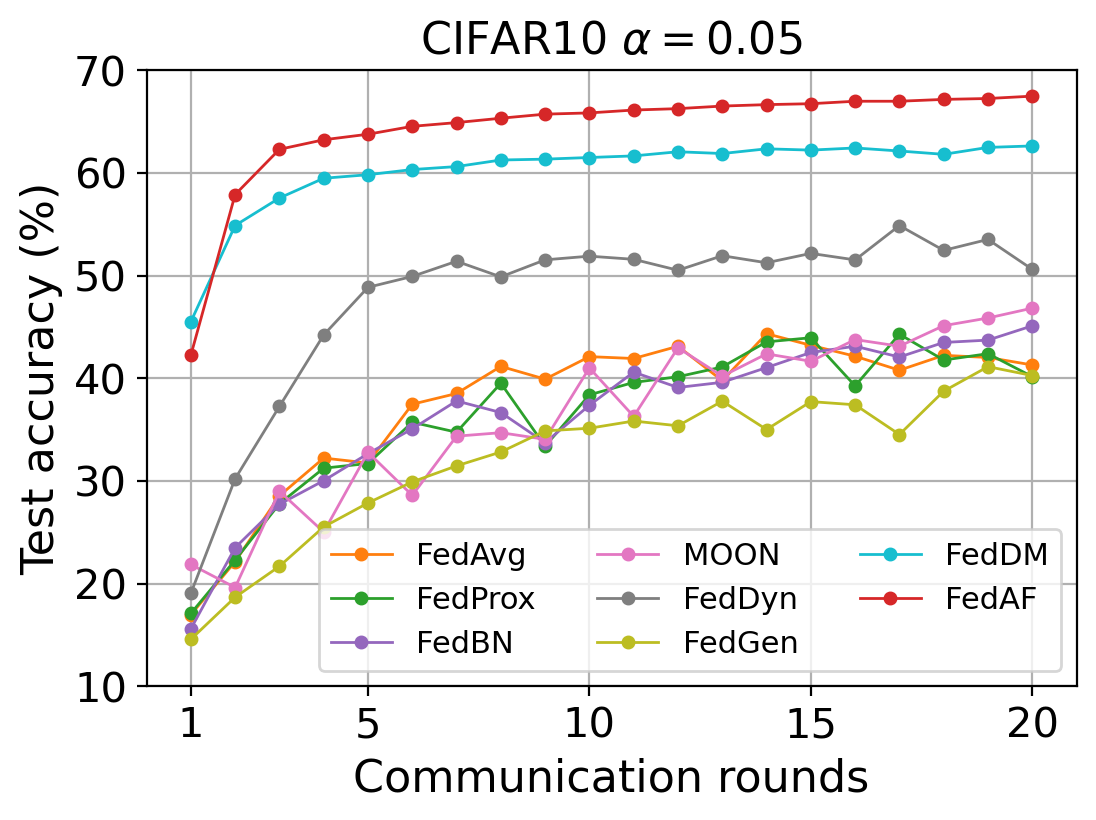}
        \caption{}
        \label{fig:result-b}
    \end{subfigure}
    \hfill
    \begin{subfigure}{0.3\linewidth}
        \includegraphics[width=1.0\textwidth]{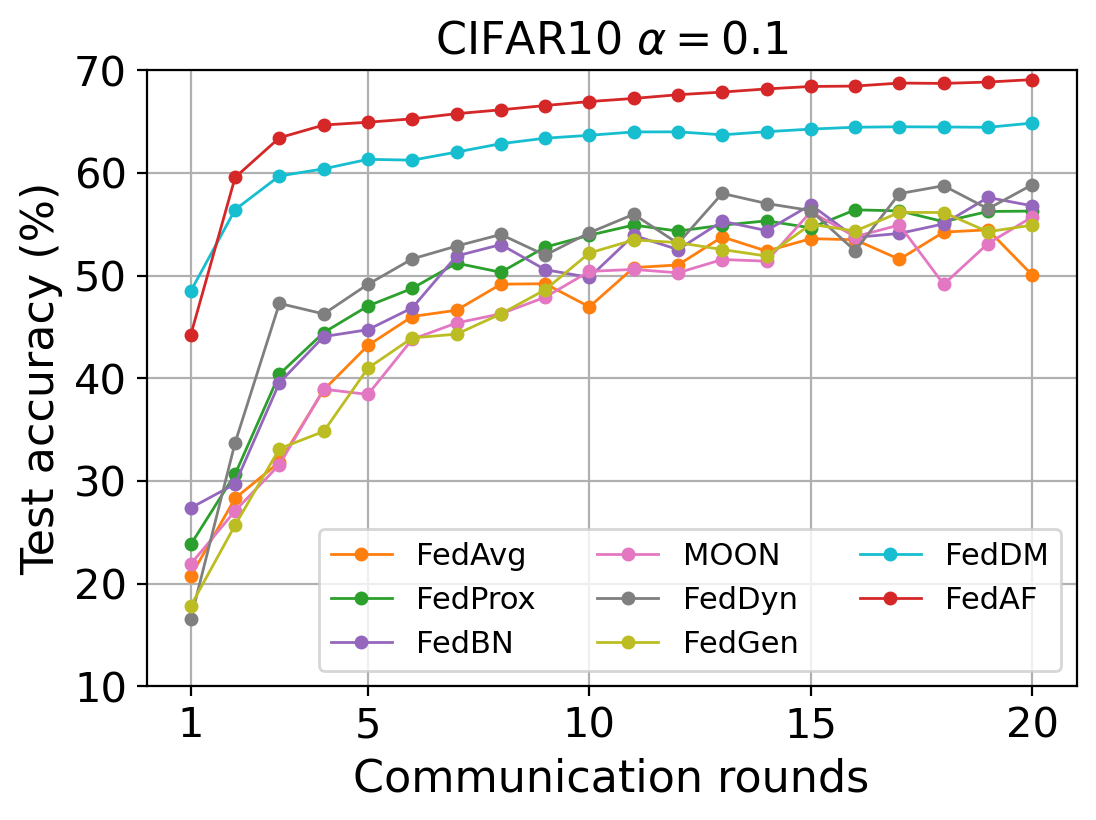}
        \caption{}
        \label{fig:result-c}
    \end{subfigure}

    \begin{subfigure}{0.3\linewidth}
        \includegraphics[width=1.0\textwidth]{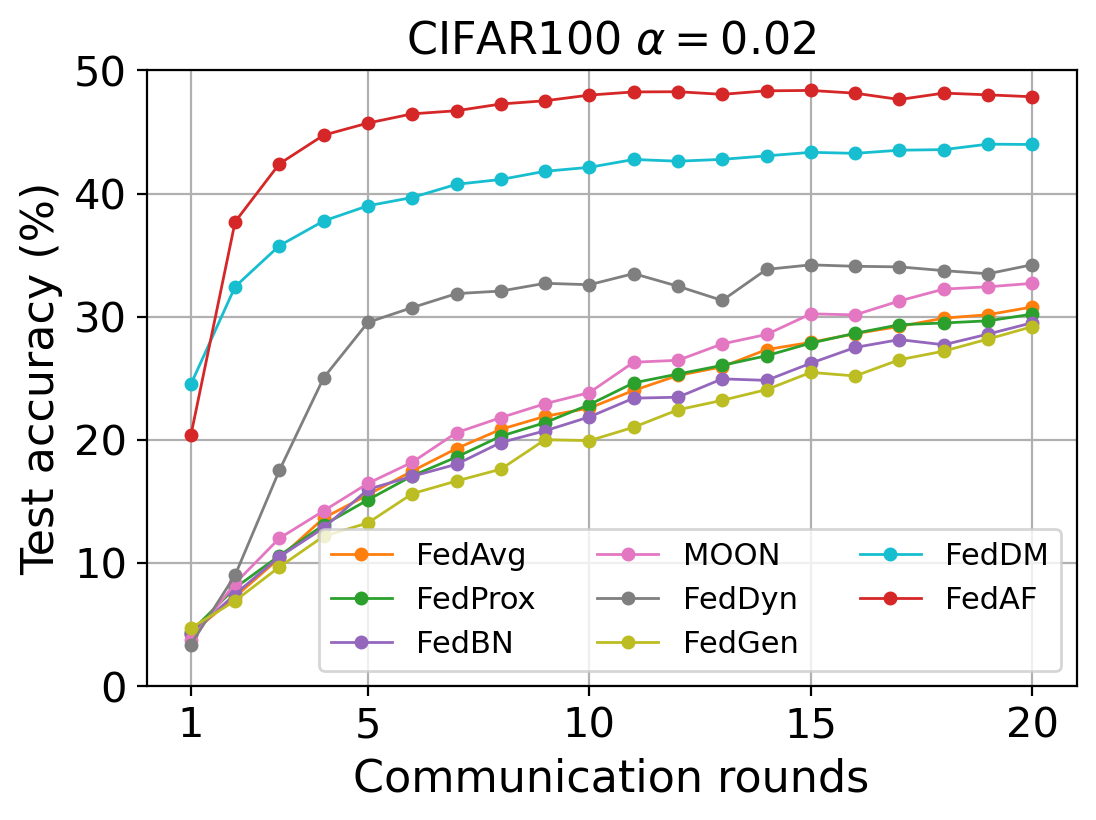}
        \caption{}
        \label{fig:result-d}
    \end{subfigure}
    \hfill
    \begin{subfigure}{0.3\linewidth}
        \includegraphics[width=1.0\textwidth]{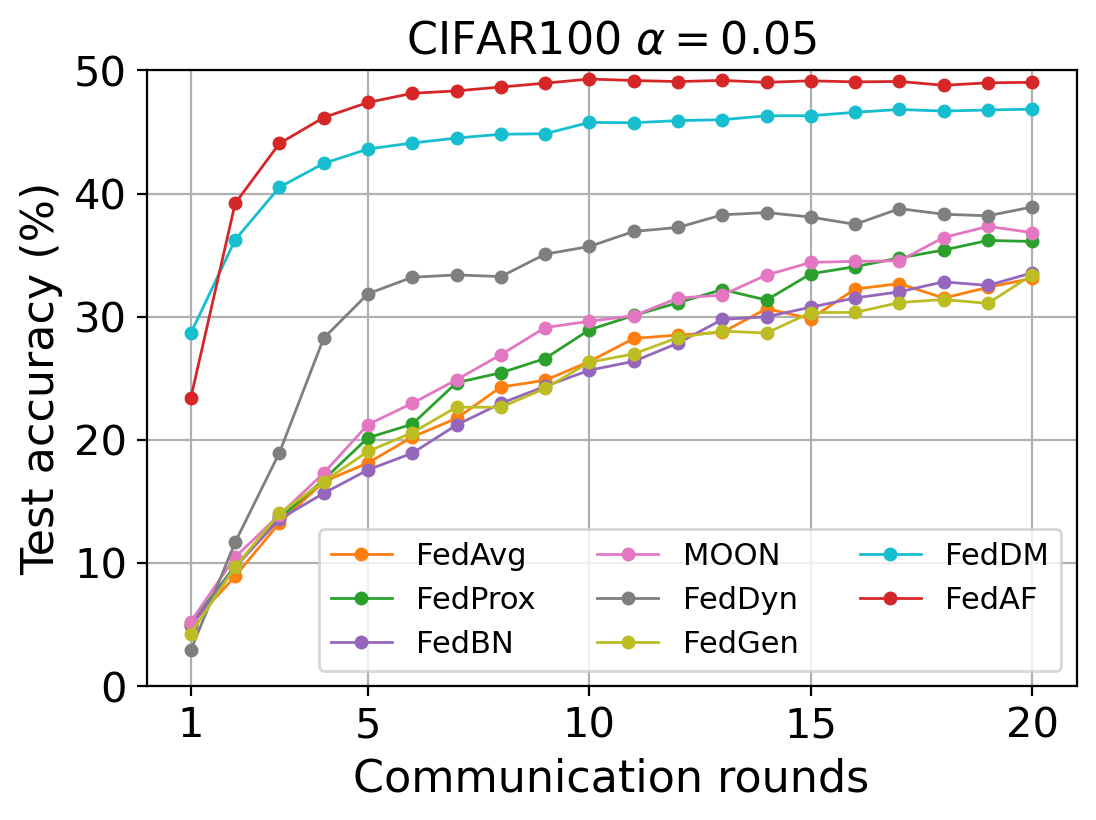}
        \caption{}
        \label{fig:result-e}
    \end{subfigure}
    \hfill
    \begin{subfigure}{0.3\linewidth}
        \includegraphics[width=1.0\textwidth]{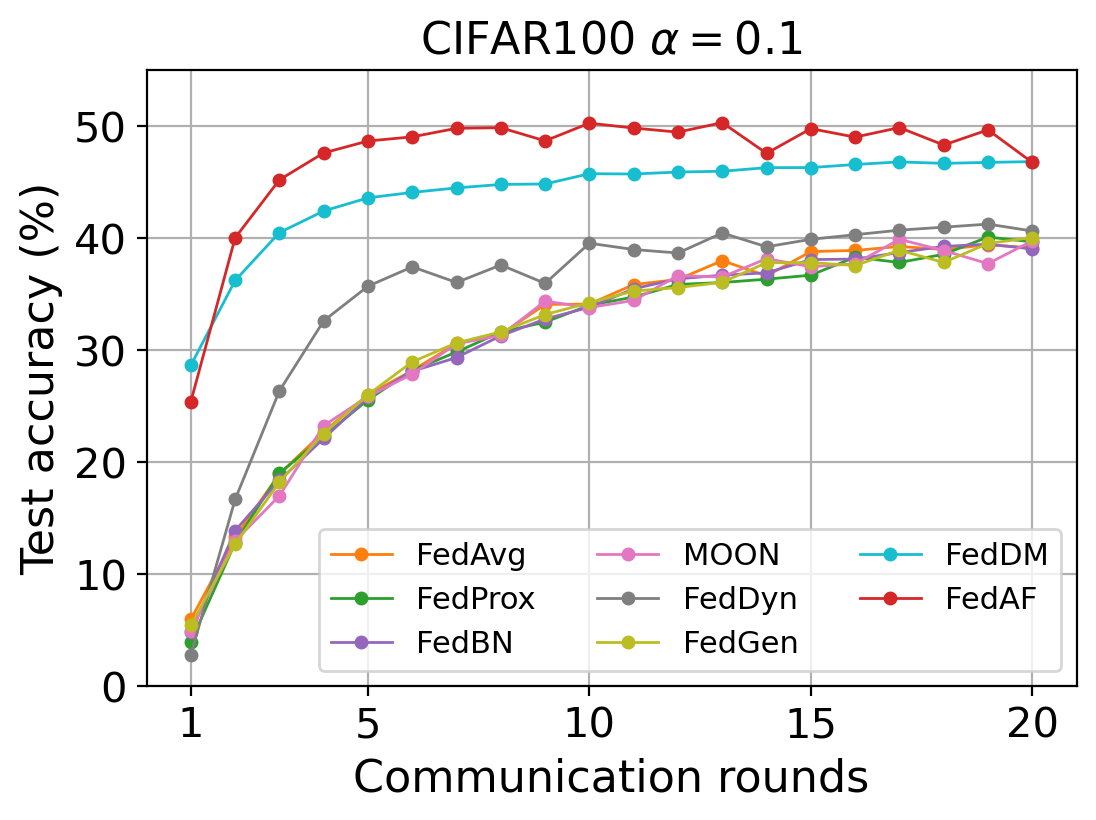}
        \caption{}
        \label{fig:result-f}
    \end{subfigure}

    \begin{subfigure}{0.3\linewidth}
        \includegraphics[width=1.0\textwidth]{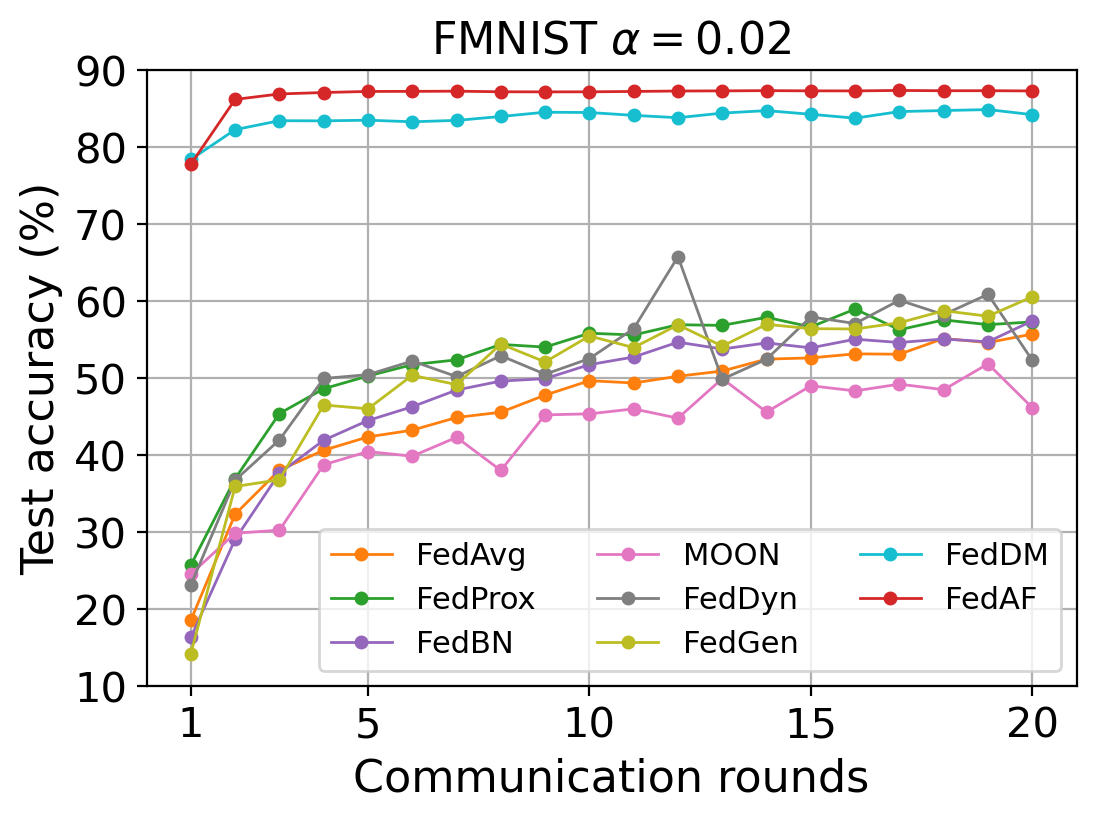}
        \caption{}
        \label{fig:result-g}
    \end{subfigure}
    \hfill
    \begin{subfigure}{0.3\linewidth}
        \includegraphics[width=1.0\textwidth]{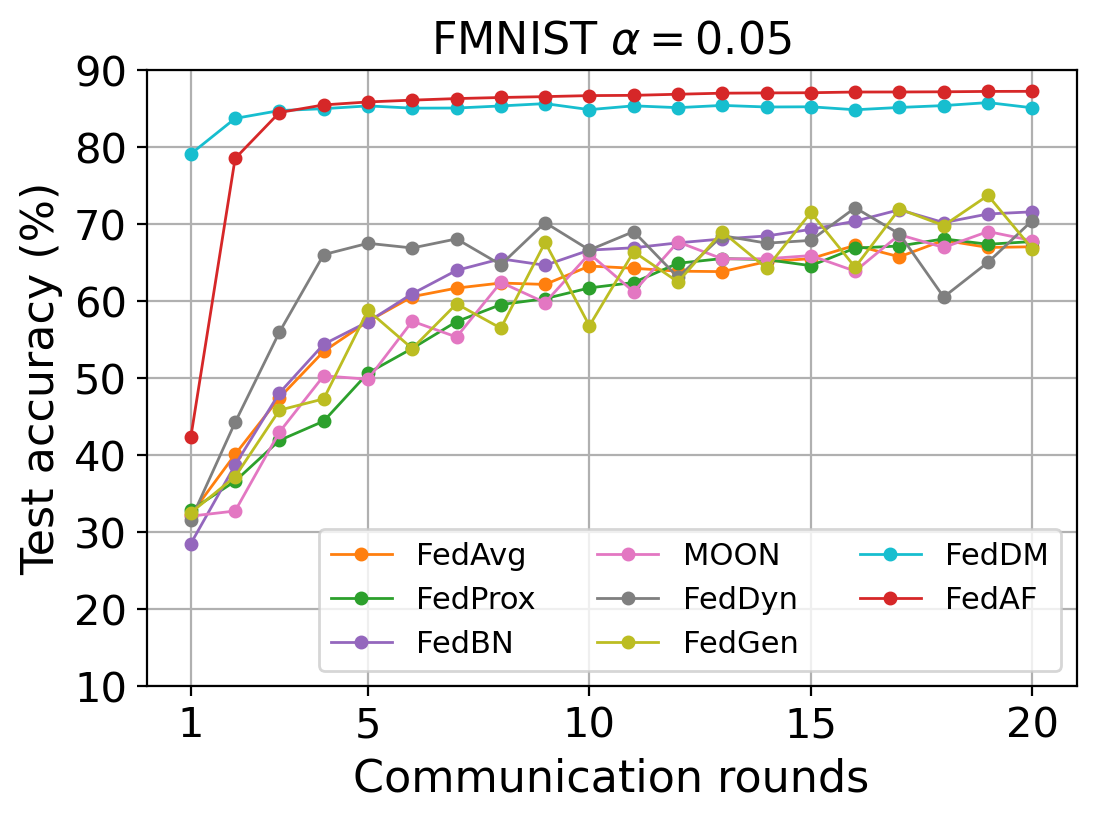}
        \caption{}
        \label{fig:result-h}
    \end{subfigure}
    \hfill
    \begin{subfigure}{0.3\linewidth}
        \includegraphics[width=1.0\textwidth]{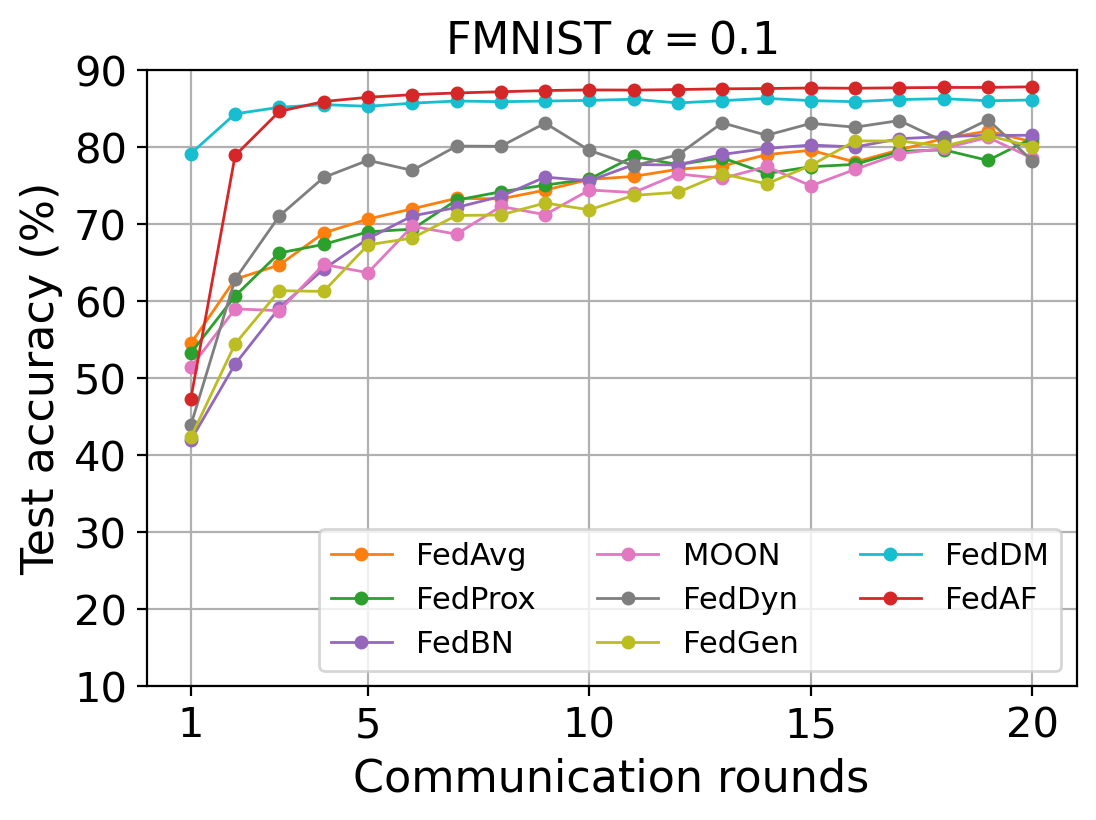}
        \caption{}
        \label{fig:result-i}
    \end{subfigure}

    \vskip -10 pt
    \caption{Comparison of convergence performance amongst baseline approaches. (a) to (c): learning curves obtained CIFAR10, (d) to (f): learning curves obtained on  CIFAR100, (g) to (i): learning curves obtained on FMNIST. In addition to the the improvement in accuracy, FedAF also stands out to deliver considerably accelerated convergence speed, especially on harder dataset.}
    \vskip -15 pt
    \label{fig:convergence-overview}
\end{figure*}

\subsection{Results for Label-skew Data Heterogeneity}
\label{ssec:results-label-noniid}
\noindent\textbf{Configuration and Hyperparameters.} 
We consider $K=10$ clients and partition the training split of each benchmark dataset into multiple data shards to simulate the local training dataset for every client. Specifically, we leverage Dirichlet distribution to partition data among clients. For every benchmark dataset, we consider three degrees of data heterogeneity, represented by $\alpha$=0.02, $\alpha$=0.05, and $\alpha$=0.1, respectively. The hyperparameter $\alpha$ controls the strength of heterogeneity. Notably, a smaller $\alpha$ implies a higher non-IID in the data distribution among clients. We choose these $\alpha$ values to simulate harsh scenarios of data heterogeneity that can be encountered in real world applications. 

For \textit{aggregate-then-adapt} baseline methods, we adopt 10 local epochs with local learning rate of 0.01 and local batch size of 64. For both FedDM and FedAF, we adopt a local batch size of 256, local update steps of 1000, 50 image-per-class (IPC) for local data condensation, while using 500 epochs with a batch size of 256 and learning rate of 0.001 for global model training. We initialize each class of condensed data using the average of randomly sampled local original data. For FedAF, the image learning rate is set to 1.0, 0.1, and 0.2, for CIFAR10, CIFAR100, and FMNIST, respectively. For FedDM, we adopt an image learning rate of 1.0 and clip the norm of gradients at 2.0 to ensure its stability during local data condensation. The global model re-sampling coefficient $\gamma$ is set to 0.9 whereas we set $\rho$=5 in FedDM (See Appendix~\ref{sec:appdix-implementation} for more implementation details).

\medskip
\noindent\textbf{Model accuracy.}
We first evaluate the highest accuracy of the global model achieved by each algorithm within 20 communication rounds. Table~\ref{tab:overview} and Figure~\ref{fig:convergence-overview} show that, FedAF significantly outperforms all \textit{aggregate-then-adapt} baselines in various settings, showing notable improvements in both mean accuracy and variance. A detailed analysis in Table~\ref{tab:overview} indicates that FedAF enhances performance compared to FedAvg by up to 25.44\%, 17.91\%, and 31.03\% on CIFAR10, CIFAR100, and FMNIST, respectively. Even against FedDyn, the leading \textit{aggregate-then-adapt} baseline, FedAF maintains an edge of up to 19.43\%, 13.70\%, and 17.74\% on the same datasets. Moreover, FedAF consistently outperforms FedDM, with accuracy advantages reaching 4.87\%, 4.56\%, and 2.17\% on CIFAR10, CIFAR100, and FMNIST, respectively. Notably, FedAF's performance is more pronounced under stronger data heterogeneity, such as at $\alpha=0.02$, demonstrating the effectiveness of our collaborative data condensation and local-global knowledge matching approaches.

\medskip
\noindent\textbf{Convergence Performance.} 
We further examine and compare the convergence speed of FedAF with baselines, the learning curves resulting from three benchmark datasests are illustrated in Figure~\ref{fig:convergence-overview}, which indicates that FedAF consistently surpasses other baseline methods in convergence speed, particularly under significant data heterogeneity. For instance, at $\alpha=0.02$, FedAF achieves the highest accuracy of other \textit{aggregate-then-adapt} baselines within just two rounds. Compared to FedDM, FedAF also maintains a similar edge; on CIFAR10 with $\alpha=0.02$, while FedDM reaches a mean accuracy of 60\% in fifteen rounds, FedAF attains this target in only three rounds, marking an 80\% increase in convergence speed.

\begin{table}[!t]
    \centering
    \setlength{\tabcolsep}{3.5pt}
    \fontsize{9pt}{9pt}\selectfont
    \begin{tabular}{l|c|c|c|c|c|c|c}
        \toprule
        \multirow{2}{*}{Methods} & \multicolumn{7}{c}{DomainNet}  \\
                & C     & I     & P     & Q     & R      & S     & Avg \\
        \midrule
        FedAvg  & 43.03          & 40.76          & 59.16          & 39.60          & 41.03           & 28.46          & 42.01\\
        FedProx & 44.81          & 43.76          & 60.22          & 38.13          & 41.55           & 29.18          & 42.94 \\
        FedBN   & 46.07          & 34.27          & 52.01          & 43.10          & 47.33           & 29.72          & 42.08 \\
        MOON    & 48.80          & 37.97          & 56.26          & 48.07          & 42.02           & 29.72          & 43.81 \\
        FedDyn  & 48.04          & \textbf{60.03} & \textbf{67.46} & 37.73          & 41.77           & 32.67          & 47.95 \\
        FedGen  & 42.77          & 37.88          & 54.37          & 37.33          & 42.86           & 25.69          & 40.15 \\
        \midrule
        FedDM   & \textbf{52.28} & 41.38          & 60.58          & \uds{62.37}    & \uds{52.45}     & \uds{46.69}    & \uds{52.62} \\
        FedAF   & \uds{51.2}     & \uds{47.05}    & \uds{62.53}    & \textbf{64.6}  & \textbf{52.64}  & \textbf{50.06} & \textbf{54.68} \\
        \bottomrule    
    \end{tabular}
    \vskip -5 pt
    \caption{Comparison of global model accuracy across various FL algorithms on the DomainNet dataset. The domains are represented by C (Clipart), I (Infograph), P (Painting), Q (Quickdraw), R (Real), and S (Sketch), ``Avg" denotes the avarage accuracy across domains. The highest and second-highest accuracies in each column are indicated by boldface and underline, respectively.}
    \vskip -10 pt
    \label{tab:domainnet}
\end{table}

\begin{figure}
\centering
\includegraphics[width=0.4\textwidth]{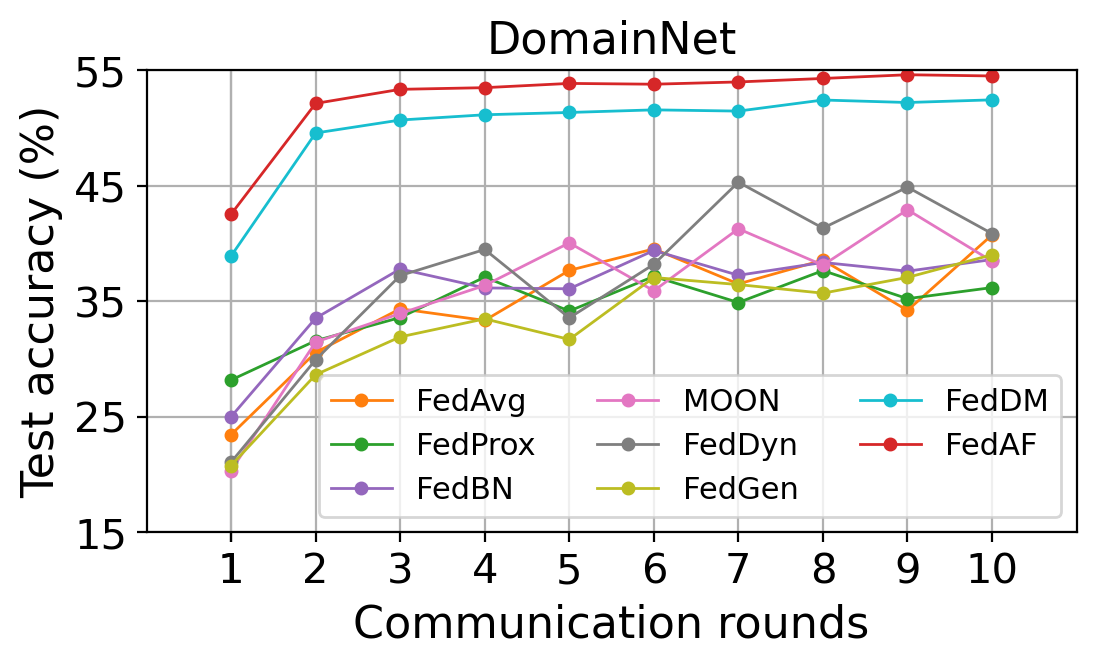}
\vskip -10 pt
\caption{Comparison of convergence performance amongst baseline approaches on DomainNet dataset. FedAF also outperform the other baselines on both accuracy and convergence in feature-skew heterogeneous data distribution.}
\vskip -8 pt
\label{fig:convergence-domainnet}
\end{figure}

\begin{table}[!t]
    \centering
    \setlength{\tabcolsep}{3.5pt}
    \fontsize{9pt}{9pt}\selectfont
    \begin{tabular}{c|c|c|c}
    \toprule
        Configuration &  $\alpha$=0.02 & $\alpha$=0.05 & $\alpha$=0.1  \\
        \midrule
        IPC=10        & 53.39$\pm$2.09 & 55.33$\pm$0.81 & 56.15$\pm$0.42  \\
        \midrule
        IPC=20        & 58.56$\pm$0.55 & 60.89$\pm$0.11 & 61.79$\pm$0.59  \\
        \midrule
        IPC=50        & 65.15$\pm$0.86 & 67.50$\pm$0.76 & 69.11$\pm$0.86  \\
        \midrule
        IPC=80        & 67.94$\pm$1.18 & 70.07$\pm$0.45 & 70.72$\pm$0.37 \\
        \midrule 
        IPC=100       & 69.14$\pm$0.56 & 71.27$\pm$0.58 & 71.66$\pm$0.37 \\
        \bottomrule
    \end{tabular}
    \vskip -5 pt
    \caption{Impact of IPC on the global model accuracy for learning CIFAR10 under three different degrees of heterogeneity.}
    \vskip -15 pt
    \label{tab:appendix-ipc}
\end{table}

\begin{figure*}
    \centering
    \begin{subfigure}{0.24\linewidth}
        \includegraphics[width=1.0\textwidth]{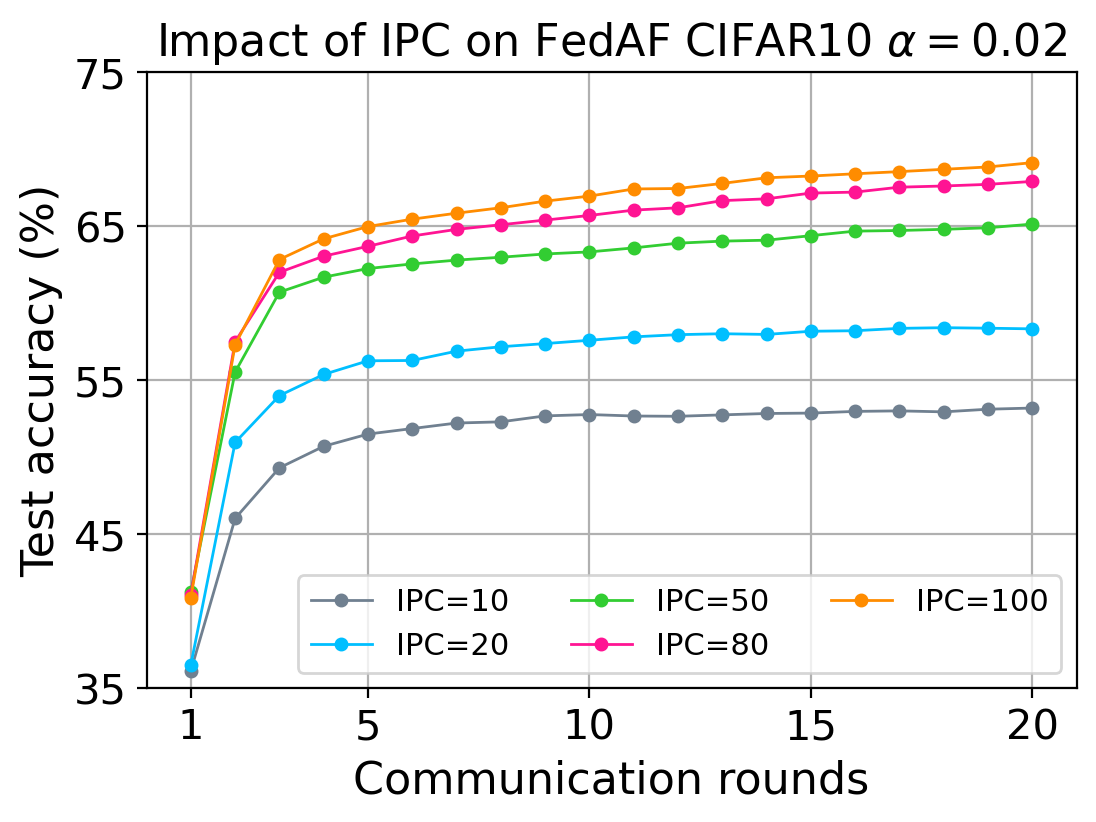}
        \caption{}
        \label{fig:ipc-cifar10-002}
    \end{subfigure}
    \hfill
    \begin{subfigure}{0.24\linewidth}
        \includegraphics[width=1.0\textwidth]{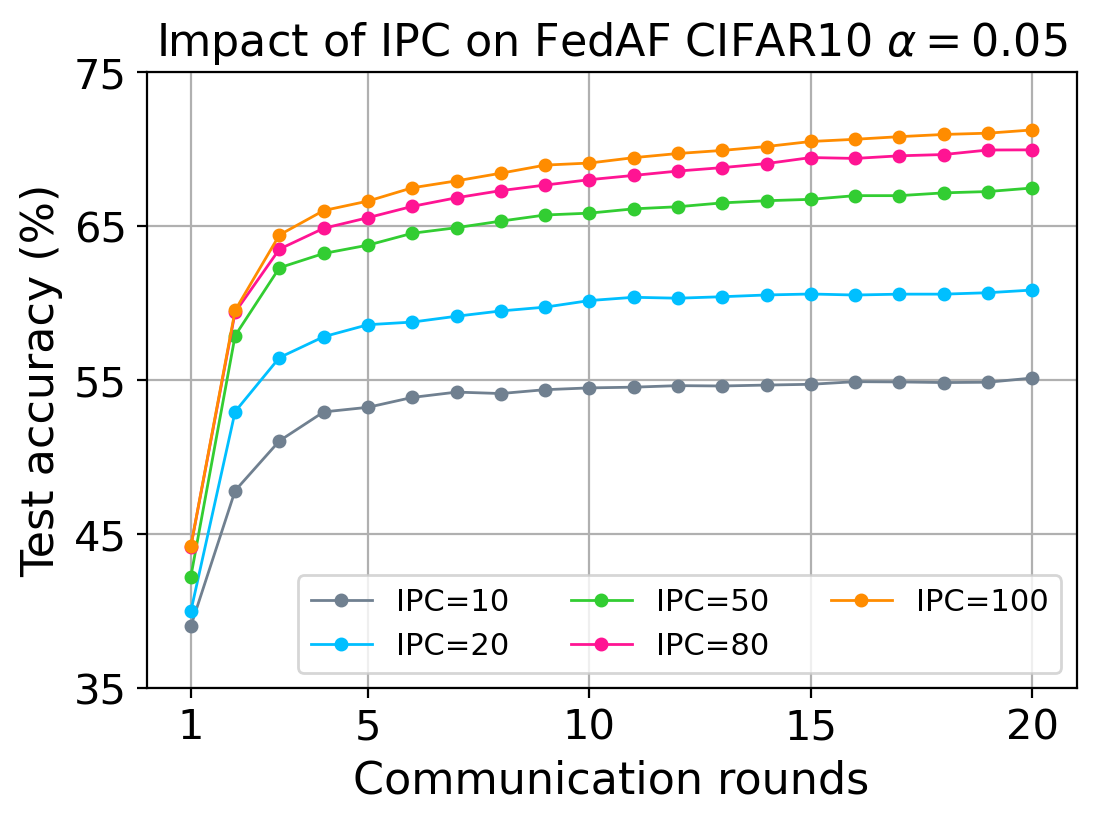}
        \caption{}
        \label{fig:ipc-cifar10-005}
    \end{subfigure}
    \hfill
    \begin{subfigure}{0.24\linewidth}
        \includegraphics[width=1.0\textwidth]{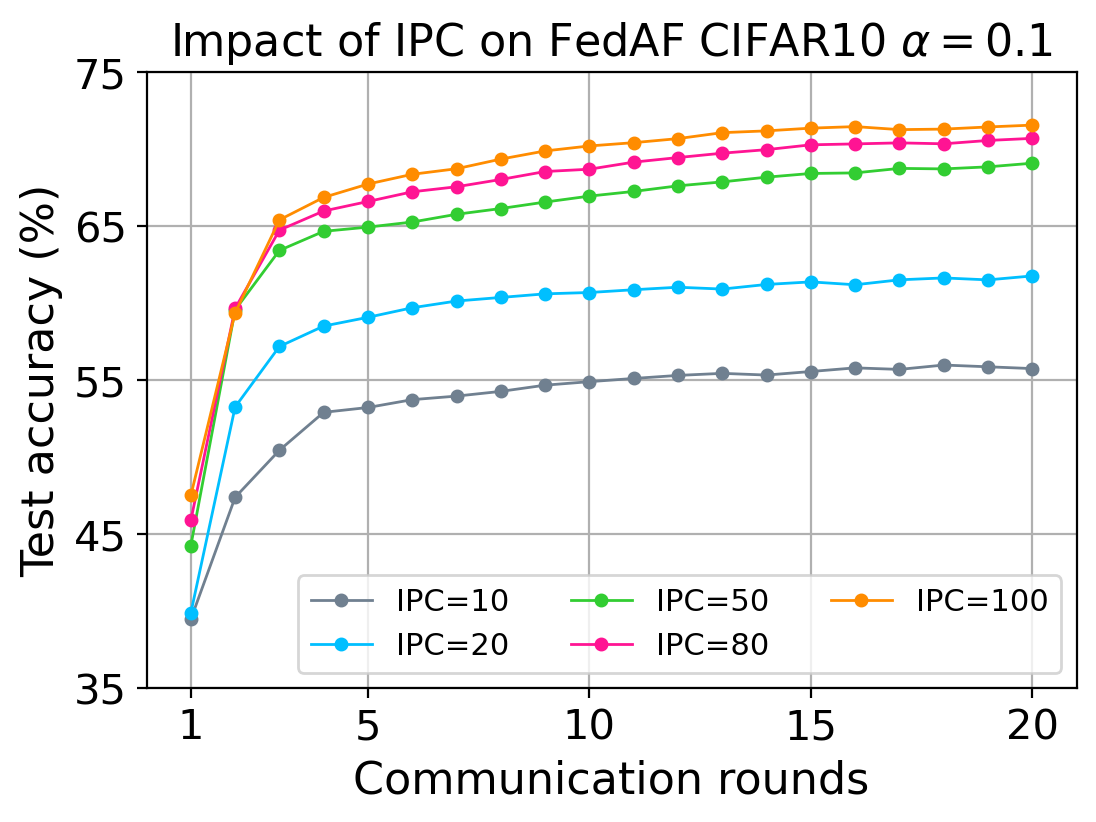}
        \caption{}
        \label{fig:ipc-cifar10-001}
    \end{subfigure}
    \hfill
    \begin{subfigure}{0.25\linewidth}
        \includegraphics[width=1.0\textwidth]{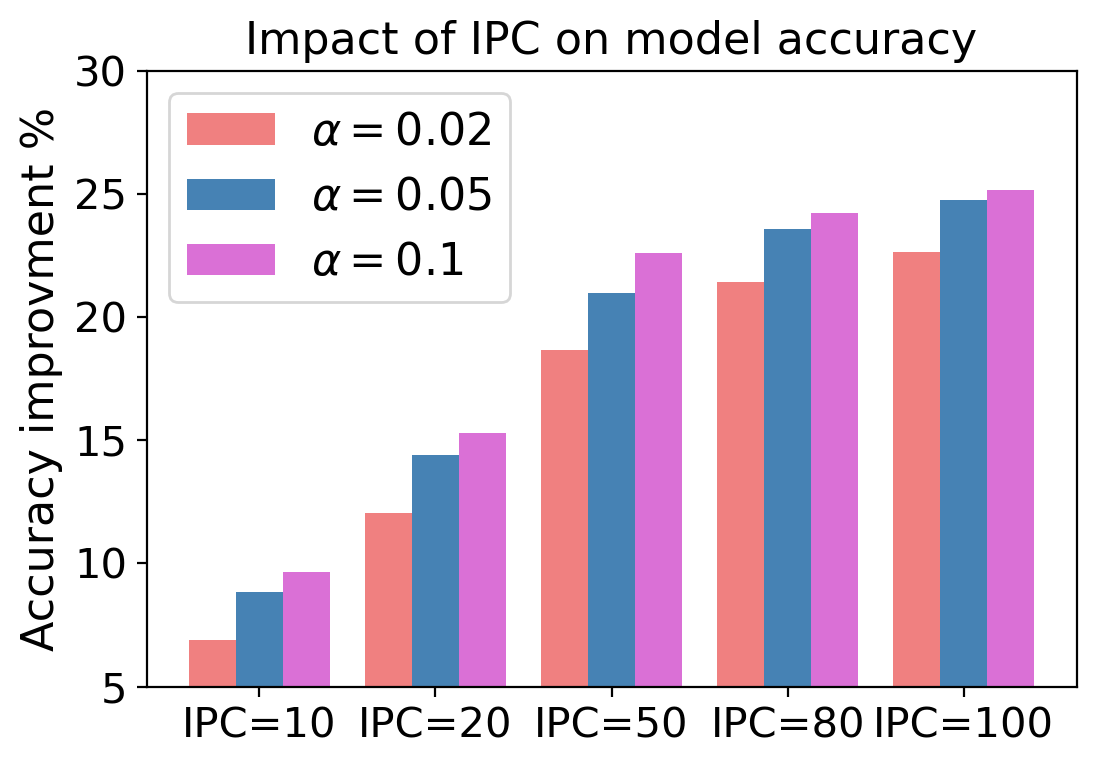}
        \caption{}
        \label{fig:ipc-cifar-gain}
    \end{subfigure}
    \vskip -10 pt
    \caption{Impact of IPC on the learning performance of FedAF on CIFAR10. (a) to (c): the resulting learing curves. (d): improvement in model accuracy compared to FedAvg. Generally, a higher IPC correlates with enhanced performance, with all tested IPC values demonstrating improvements over FedAvg.}
    \vskip -15 pt
    \label{fig:ablation-ipc}
\end{figure*}

\subsection{Result for Feature-skew Data Heterogeneity}
\label{ssec:result-feature-noniid}
\noindent\textbf{Configuration and Hyperparameters.} For the experiments in feature-skew scenario, we follow \cite{li:21fedbn} to form a sub-dataset of DomainNet comprising only the top ten most frequent classes across all domains. We configure six clients, each holding data from a unique domain, to mimic real-world scenarios such as different hospitals using distinct imaging protocol and equipment, influencing the data heterogeneity. All algorithms are run for ten communication rounds to compare the resulting global model's accuracy for every domain and the average accuracy across domains. For \textit{aggregate-then-adapt} baselines, we maintain the same hyperparameters as in the label-skew scenarios. Both FedDM and FedAF use an image learning rate of 1.0, with the other settings such as batch size and number of local steps consistent with those in the label-skew scenarios (see Appendix~\ref{sec:appdix-implementation} for more details).   

\medskip
\noindent\textbf{Model Accuracy and Convergence Performance.} From Table~\ref{tab:domainnet} and Figure~\ref{fig:convergence-domainnet}, it is evident that FedAF outperforms all other baseline methods in average accuracy and related convergence performance across all domains. This comparison highlights FedAF's consistent superiority, ranking as either the top or the second-best performer in every domain. Similar to the label-skew scenarios, \textit{aggregate-then-adapt} FL approaches are found to be less effective compared to FedAF. Moreover, FedAF not only demonstrates higher accuracy but also exhibits faster convergence performance than FedDM. For instance, FedAF achieves the accuracy level in just two rounds that FedDM requires ten rounds to reach, indicating an 80\% acceleration. These advantages underscore the benefit of integrating knowledge from other domains through our collaborative data condensation and local-global knowledge matching strategies, validating the effectiveness of these approaches.

\subsection{Performance Analysis of FedAF}
\noindent\textbf{Impact of IPC.}
We conduct an ablation study on CIFAR10 to examine how variations in IPC might potentially influence the performance of FedAF under various degrees of data heterogeneity. We summarize the results in Table~\ref{tab:appendix-ipc}, and also illustrate and compare the resulting learning performance in Figure~\ref{fig:ablation-ipc}. It can be concluded that higher IPC values generally lead to higher accuracy. Besides, for each IPC value, the resulting model accuracy does not vary significantly over different values of $\alpha$ (with the gap between the highest and the lowest being 3.96\%). From Figure~\ref{fig:ipc-cifar-gain} we find that although remarkable performance gain can be observed when IPC ranges from 10 to 50, this improvement starts to become marginal using an even higher IPC. Considering the increase in communication cost is approximately linear with the value of IPC. On the other hand, the lower the compression ratio (i.e., the ratio between the amount of condensed and original data), the higher the privacy retention \cite{dong22:dcprivacy}. Therefore, an IPC value lower than 50 should be generally considered as a trade-off between performance, communication cost, and privacy. 

\begin{table}[!t]
    \centering
    \setlength{\tabcolsep}{3.5pt}
    \fontsize{9pt}{9pt}\selectfont
    \begin{tabular}{c|c|c|c}
    \toprule
        Configuration &  $\alpha$=0.02 & $\alpha$=0.05 & $\alpha$=0.1  \\
        \midrule
        FedAF     & 65.15$\pm$0.86 & 67.50$\pm$0.76 & 69.11$\pm$0.86  \\
        \midrule
        w/o CDC  & 64.16$\pm$0.83 & 65.88$\pm$0.93 & 67.90$\pm$0.53  \\
        \midrule
        w/o LGKM & 64.12$\pm$0.85 & 66.27$\pm$1.31 & 68.14$\pm$0.81  \\
        \midrule
        FedDM   & 60.28$\pm$0.82 & 62.97$\pm$0.96 & 64.88$\pm$0.35 \\
        \bottomrule
    \end{tabular}
    \vskip -5 pt
    \caption{Impact of core design on the global model accuracy for learning CIFAR10 under three different degrees of heterogeneity. ``w/o CDC'' denotes the FedAF without collaborative data condensation, ``w/o LGKM'' denotes the FedAF without the local-global knowledge matching.}
    \vskip -15 pt
    \label{tab:core_design}
\end{table}

\medskip
\noindent\textbf{Impact of Core Designs.} 
The collaborative data condensation and local-global knowledge matching act as two fundamental techniques of FedAF. We conduct additional experiments on CIFAR10 to further reveal how these two techniques contribute to performance improvement. Specifically, we compare the full FedAF with two other configurations where each fundamental technique is not utilized. We also compare it with FedDM, where none of these techniques exist. As shown in Table~\ref{tab:core_design}, the mean accuracy resulting from using just one of the fundamental techniques alone still shows noticeable improvement over that obtained with FedDM. Moreover, the full FedAF exhibits further improvement in the mean accuracy. These results verify that by promoting the utilization of additional knowledge derived from data distributed across clients, FedAF indeed can empower clients to learn higher quality condensed data and train the global model with improved performance, which eventually contributes to the enhancement in the overall learning performance.

\medskip
\noindent\textbf{Impact of Model Re-sampling.}
As the value of model re-sampling coefficient $\gamma$ in \eqref{eq:model_resample} controls the interpolation between the parameters of the global model received from the server and those parameters randomly sampled, we analyze the effect of this technique by running ten rounds of FedAF on CIFAR10 with $\alpha$=0.1 and comparing the resulting accuracy over different choices of $\gamma$. 
\begin{table}
    \centering
    \setlength{\tabcolsep}{3.5pt}
    \fontsize{9pt}{9pt}\selectfont
    \begin{tabular}{c|c|c|c|c|c}
    \toprule
     $\gamma$   & 0.2       & 0.5         & 0.8     & 0.9     & 1.0 \\
     \midrule
     Accuracy   & 61.30    & 63.52        & 66.15   & 66.97  & 64.92 \\
     \bottomrule
    \end{tabular}
    \vskip -5 pt
    \caption{The influence of global model re-sampling on learning performance. The first row lists the values of $\gamma$ tested and the second row reports the corresponding accuracy.}
    \vskip -15 pt
    \label{tab:tab-model-resample}
\end{table}
Note that a larger $\gamma$ indicates the re-sampled model tends to retain more knowledge learned in the global model from the previous round, while $\gamma$=0 implies clients use a model with randomly initialized parameters for data condensation. As expected, Table~\ref{tab:tab-model-resample} shows that larger $\gamma$ values generally lead to higher model accuracy. However, the optimal $\gamma$ is found to be 0.9, rather than the maximal possible value of 1.0, verifying that a suitable random perturbation to the true global model can indeed regulate the data condensation and emhance learning performance. Notably, $\gamma$=0.8 also yields comparable accuracy, suggesting that FedAF's performance is robust to variations in $\gamma$. Additionally, we experimented with $\gamma$=0. However, using purely random weight parameters, the model struggled to stabilize the data condensation process, leading to its exclusion from the further comparison.

\section{Conclusion}
This paper presents FedAF, a novel FL algorithm designed to tackle data heterogeneity with an aggregation-free framework. FedAF grants clients and the server richer insights about the original data distributed across clients through collaborative data condensation and local-global knowledge matching. This strategy effectively addresses cross-client data heterogeneity and boosts the learning performance of both condensed data and the global model. Consequently, FedAF demonstrates considerable improvement over state-of-the-art FL methods in terms of model accuracy and convergence speed. 

\medskip
{\noindent
\textbf{Acknowledgement.} 
This work is supported by the Agency for Science, Technology and Research (A*STAR) under its IAF-ICP Programme (Award No: I2301E0020). 
This work is also supported by the National Research Foundation, Singapore under its AI Singapore Programme (Award No: AISG2-TC-2021-003).
This work is also partially supported by A*STAR Central Research Fund ``A Secure and Privacy Preserving AI Platform for Digital Health”.
This work is also supported by A*STAR Career Development Fund (No. C222812010).
}

\clearpage
{
    \small
    \bibliographystyle{ieeenat_fullname}
    \bibliography{main}
}

\clearpage
\setcounter{page}{1}
\appendix

\begin{figure*}[!t]
    \centering
    \begin{subfigure}{0.25\linewidth}
        \includegraphics[width=1.0\textwidth]{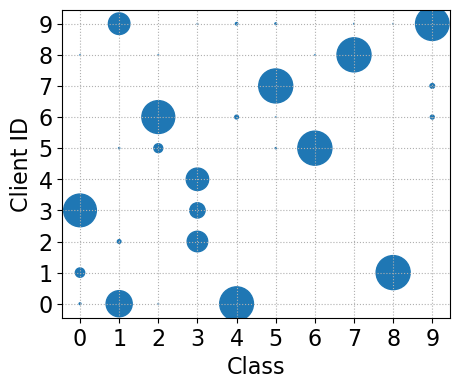}
        \caption{CIFAR10 $\alpha$=0.02}
        \label{fig:suppl-cifar10-002}
    \end{subfigure}
    \begin{subfigure}{0.25\linewidth}
        \includegraphics[width=1.0\textwidth]{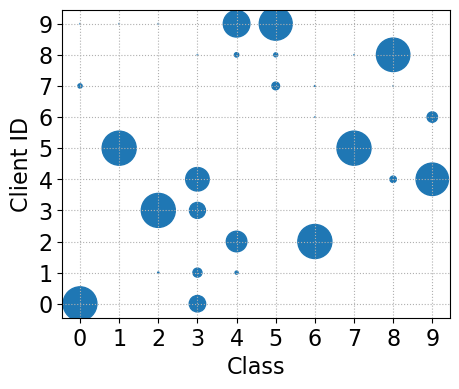}
        \caption{CIFAR10 $\alpha$=0.05}
        \label{fig:suppl-cifar10-005}
    \end{subfigure}
    \begin{subfigure}{0.25\linewidth}
        \includegraphics[width=1.0\textwidth]{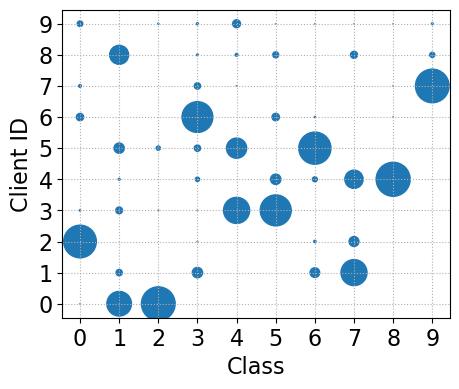}
        \caption{CIFAR10 $\alpha$=0.1}
        \label{fig:suppl-cifar10-001}
    \end{subfigure}
    \vskip -5 pt
    \caption{Visualization of cross-client data distribution for CIFAR10 dataset under three different degrees of label-skew heterogeneity.}
    \vskip -10 pt
    \label{fig:suppl-data-distribution-cifar10}
\end{figure*}
\begin{figure*}[ht]
    \centering
    \begin{subfigure}{0.25\linewidth}
        \includegraphics[width=1.0\textwidth]{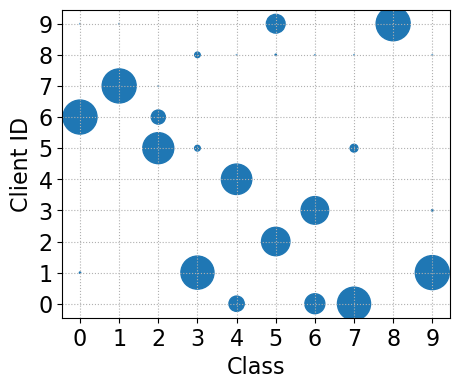}
        \caption{FMNIST $\alpha$=0.02}
        \label{fig:suppl-fmnist-002}
    \end{subfigure}
    \begin{subfigure}{0.25\linewidth}
        \includegraphics[width=1.0\textwidth]{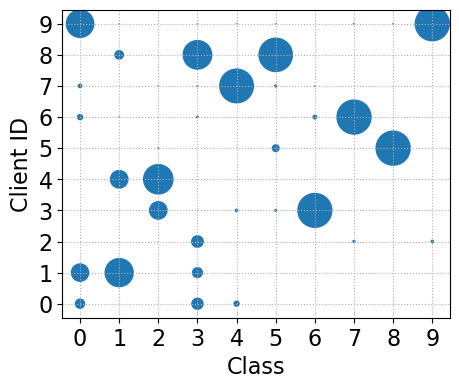}
        \caption{FMNIST $\alpha$=0.05}
        \label{fig:suppl-fmnist-005}
    \end{subfigure}
    \begin{subfigure}{0.25\linewidth}
        \includegraphics[width=1.0\textwidth]{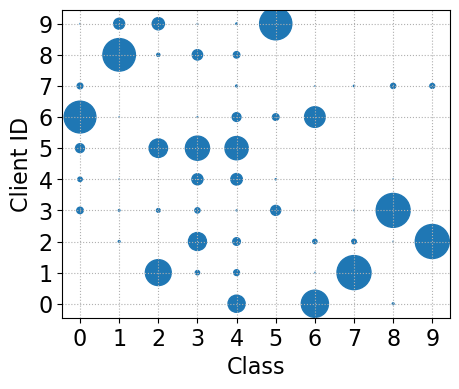}
        \caption{FMNIST $\alpha$=0.1}
        \label{fig:suppl-fmnist-001}
    \end{subfigure}
    \vskip -5 pt
    \caption{Visualization of cross-client data distribution for FMNIST dataset under three different degrees of label-skew heterogeneity.}
    \vskip -10 pt
    \label{fig:suppl-data-distribution-fmnist}
\end{figure*}

\section{Implementation Details} 
\label{sec:appdix-implementation}
\noindent\textbf{Computing Platform.}
In this paper, we use PyTorch to implement all the algorithms and experiments in both the main paper and this supplementary material. We run our experiments using an Nvidia RTX3090 GPU with 24 GB of memory. 

\medskip
\noindent\textbf{Model Architecture.}
In Section~\ref{sec:results}, we adopt a convolutional neural network (ConvNet) that follows the same architecture as reported in \cite{zhao:23dm}. The encoder of this model consists of three convolutional layers, each followed by a ReLU activation function and average pooling. To facilitate comparison with FedBN, batch normalization is incorporated into the model. A fully-connected layer serves as the classifier and is attached on top of the encoder.

\medskip
\noindent\textbf{Training Details.}
In addition to the configurations and hyperparameters detailed in Sections~\ref{ssec:results-label-noniid} and \ref{ssec:result-feature-noniid}, we have set specific values for other algorithms. For FedProx and MOON, the hyperparameter $\mu$ is set to 0.001 and 1.0, respectively. In FedDyn, we use 0.01 for the hyperparameter $\alpha$. For experiments on the DomainNet dataset with FedDM and FedAF, we employ an image-per-class (IPC) of 20, to achieve a similar condensation ratio as label-skew scenarios. Additionally, we resize the images of DomainNet into 64$\times$64 resolution. 
For FedAF, the regularization weights $(\lambda_\text{loc},\lambda_\text{glob})$ for collaborative data condensation and local-global knowledge matching are set to (0.0001,0.01), (0.0001,0.01), and (0.001, 2.0) on CIFAR10, CIFAR100, and FMNIST, respectively. On DomainNet, we use (0.01, 0.1) for $(\lambda_\text{loc},\lambda_\text{glob})$. Furthermore, for all algorithms, we employ PyTorch's built-in SGD optimizer with a momentum of 0.9. The same implementation is also used for the optimizer of local data condensation in FedDM and FedAF. The accuracy evaluation for all algorithms is conducted over the testing split of each benchmark dataset, to simulate centralized validation or test data at the server. For all experiments, we average the accuracy and learning curves over three trial runs, each with a different random seed.

\begin{figure*}
    \centering
    \begin{subfigure}{0.99\linewidth}
        \includegraphics[width=1.0\textwidth]{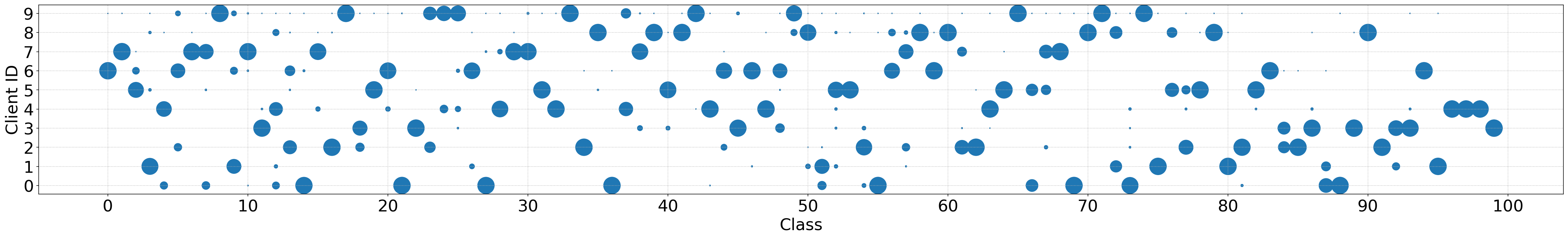}
        \caption{CIFAR100 $\alpha$=0.02}
        \label{fig:suppl-cifar100-002}
    \end{subfigure}
    \hfill
    \begin{subfigure}{0.99\linewidth}
        \includegraphics[width=1.0\textwidth]{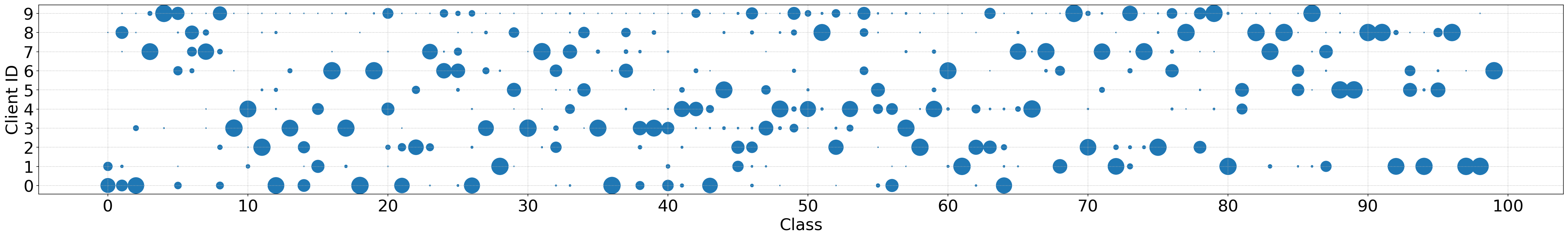}
        \caption{CIFAR100 $\alpha$=0.05}
        \label{fig:suppl-cifar100-005}
    \end{subfigure}
    \hfill
    \begin{subfigure}{0.99\linewidth}
        \includegraphics[width=1.0\textwidth]{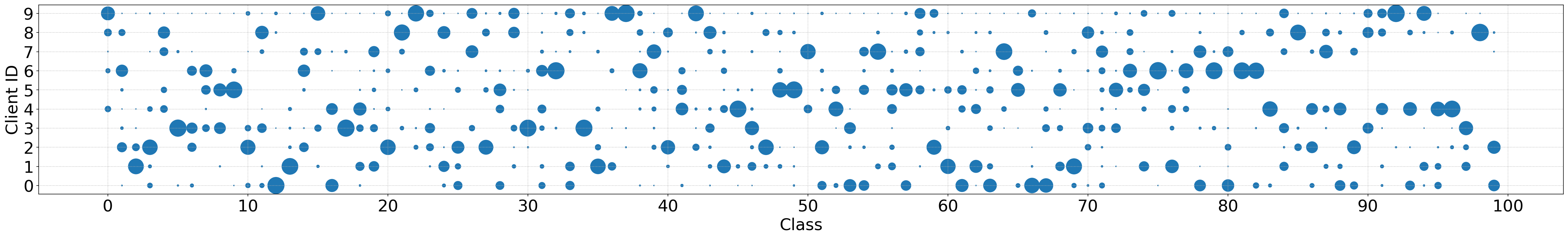}
        \caption{CIFAR100 $\alpha$=0.1}
        \label{fig:suppl-cifar100-001}
    \end{subfigure}
    \vskip -5 pt
    \caption{Visualization of cross-client data distribution for CIFAR100 dataset under three different degrees of label-skew heterogeneity.}
    \vskip -5 pt
    \label{fig:suppl-data-distribution-cifar100}
\end{figure*}

\begin{figure*}
    \centering
    \includegraphics[width=1.0\textwidth]{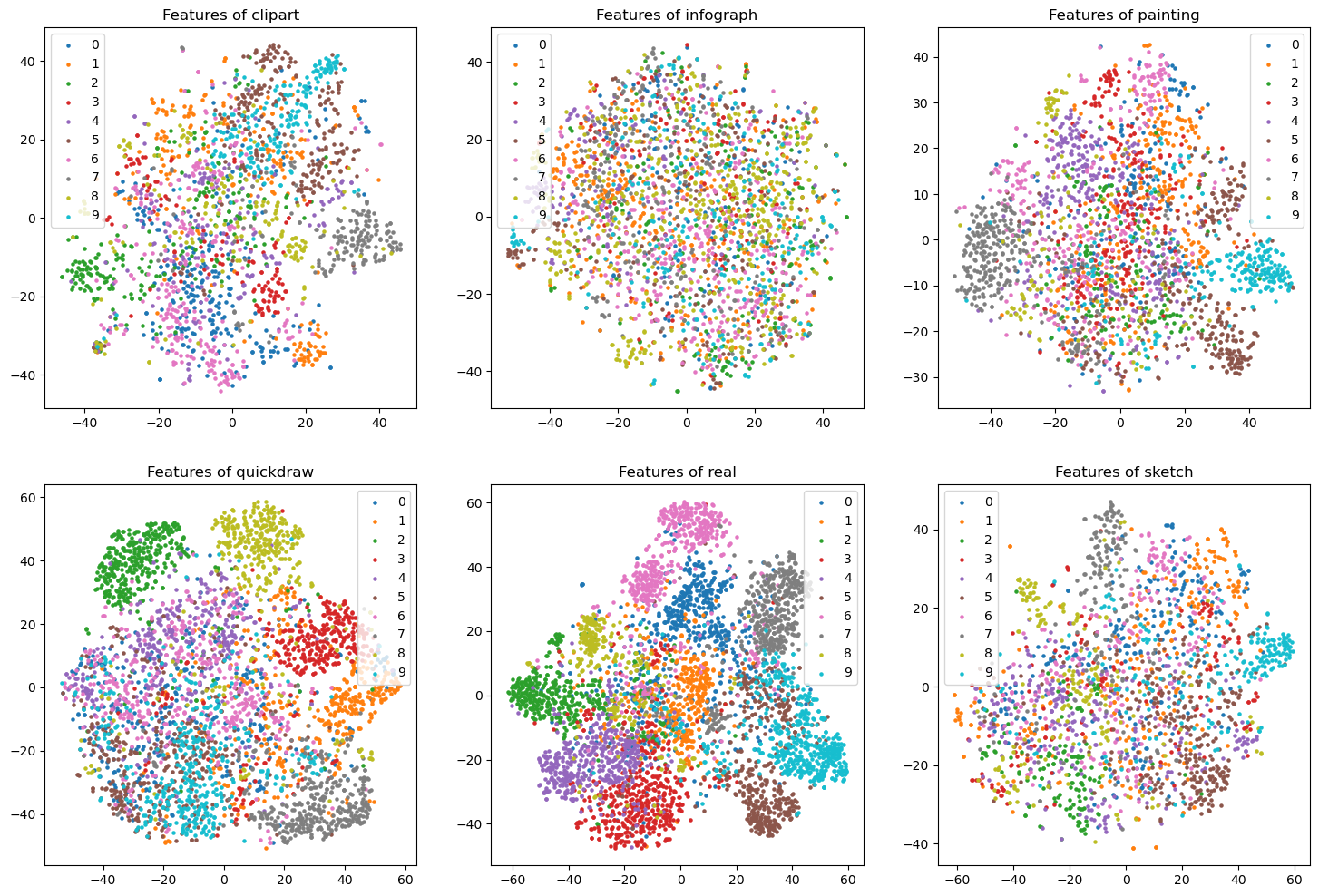}
    \caption{T-SNE visualization of features extracted from DomainNet data, using a sub-dataset split from \cite{li:21fedbn}.}
    \label{fig:suppl-tsne-domainnet}
\end{figure*}

\medskip
\noindent\textbf{Visualizing Data Distribution.} 
For label-skew data heterogeneity, we explore three degrees of non-IID in cross-client data distribution, represented by $\alpha$ values of 0.02, 0.05, and 0.1, where a smaller $\alpha$ indicating stronger heterogeneity. Figures~\ref{fig:suppl-data-distribution-cifar10}, \ref{fig:suppl-data-distribution-fmnist}, and \ref{fig:suppl-data-distribution-cifar100} show the class-wise data distribution per client for CIFAR10, CIFAR100, and FMNIST datasets, respectively, using a random seed from our experiments. In these figures, the size of the blue circles corresponds to the number of data samples. We observe that with a smaller $\alpha$, clients tend to possess data concentrated in fewer classes and share fewer common classes, indicating a more pronounced label-skew non-IID distribution. For feature-skew heterogeneity, we analyze feature distribution in Figure~\ref{fig:suppl-tsne-domainnet}. Here, domain features are extracted using the encoder of a ResNet50 \cite{he2016deep} pre-trained on ImageNet-1K \cite{deng2009imagenet}. The features are then fitted into a 2D space using t-SNE \cite{maaten2008visualizing} for visualization. Figure~\ref{fig:suppl-tsne-domainnet} clearly shows that, despite identical classes across six domains, the feature distribution of each clas varies significantly from one domain to another.

\section{More Experiment Results with ResNet18}
\label{sec:suppl-result-resnet}
In further experimentation, we evaluate the performance of FedAF and compare it to baseline methods using a ResNet18 model \cite{he2016deep} on CIFAR10 dataset. Conducting 20 communication rounds for all algorithms, the resulting accuracy are presented in Table~\ref{tab:suppl-resnet-cifar10}. As expected, FedAF consistently outperforms the baseline methods in both accuracy and related standard deviation across three different degrees of data heterogeneity. This advantage is particularly pronounced under strong heterogeneity, such as at $\alpha$=0.02, where FedAF achieves a 14.62\% higher accuracy than FedAvg and an 11.51\% improvement over MOON, the top performer of \textit{aggregate-then-adapt} baselines. Additionally, FedAF maintains steady accuracy advantages of 2\% over FedDM throughout all the $\alpha$ settings. 

\begin{table}[!t]
    \centering
    \setlength{\tabcolsep}{3.5pt}
    \fontsize{9pt}{10pt}\selectfont
    \begin{tabular}{l|c|c|c}
    \toprule
        Methods      &  $\alpha$=0.02 & $\alpha$=0.05  & $\alpha$=0.1  \\
        \midrule
        FedAvg       & 26.48$\pm$0.58 & 32.72$\pm$2.47 & 35.85$\pm$3.73  \\
        FedProx      & 26.86$\pm$2.69 & 32.73$\pm$2.45 & 36.25$\pm$2.96  \\
        FedBN        & 27.00$\pm$2.49 & 30.29$\pm$3.38 & 35.48$\pm$3.45  \\
        MOON         & 29.59$\pm$3.57 & 33.11$\pm$3.74 & 37.26$\pm$2.66 \\
        FedDyn       & 22.67$\pm$1.54 & 29.89$\pm$4.48 & 35.38$\pm$1.56 \\
        FedGen       & 26.63$\pm$2.07 & 32.48$\pm$3.04 & 38.85$\pm$2.00 \\
        \midrule 
        FedDM        & 39.18$\pm$0.29 & 39.47$\pm$0.66 & 40.83$\pm$0.67 \\
        FedAF        & \textbf{41.10}$\pm$\textbf{0.50} & \textbf{41.40}$\pm$\textbf{0.66} & \textbf{42.93}$\pm$\textbf{0.29} \\
        \bottomrule
    \end{tabular}
    \vskip -5 pt
    \caption{Comparison of accuracy achieved by various state-of-the-art baselines by training ResNet18 on CIFAR10. Three different degrees of label-skew data heterogeneity are implemented.}
    \vskip -10 pt
    \label{tab:suppl-resnet-cifar10}
\end{table}

\section{Communication Cost Analysis}
\noindent\textbf{Baseline Methods and Model Size.}
For typical \textit{aggregate-then-adapt} baseline methods like FedAvg, only the model parameters are communicated back and forth between clients and the server. In our experiments, the ConvNet and ResNet18 model has 381,450 and 11,181,642 parameters, respectively. With float32 precision, each parameter takes 4 bytes, so that the size of these two models is evaluated at 1.46 MB, and 42.65 MB, respectively.   

\medskip
\noindent\textbf{FedAF and Size of Condensed Data.}
In FedAF's upstream communication, each client $k$ sends three items to the server: 1) the local condensed data $\mathcal{S}_k$, 2) the class-wise mean logit $\mathcal{V}_k$, and 3) the class-wise mean soft labels $\mathcal{R}_k$, whereas in the downstream communication, each client $k$ downloads two items: 1) the global model $\mathbf{w}$ which shares the same architecture as that in \textit{aggregate-then-adapt} baselines, and 2) the class-wise mean logits from all other clients, denoted by $\mathcal{V}$ in \eqref{eq:collective_mean_logits}. The matrices $\mathcal{V}_k$ and $\mathcal{R}_k$ share the same size, for ten-class datasets like CIFAR10, FMNIST, and the sub-dataset we extracted from the DomainNet, both $\mathcal{V}_k$ and $\mathcal{R}_k$ include ten vectors with ten values in float32, making the size of them is approximately 4$\times$10$^{-4}$ MB each. Whereas for CIFAR100 that contains data of 100 classes, the size of $\mathcal{V}_k$ and $\mathcal{R}_k$ altogether is then evaluated at approximately 0.076 MB. Assuming that the condensed data is stored and transmitted in the PIL format so that one can use 8-bit unsigned integer (or 1 byte) for each pixel per channel, the size of every ten such condensed data samples from FMNIST is evaluated at 7.5$\times$10$^{-3}$ MB. Similarly, the size of every ten condensed data learned from CIFAR10 or CIFAR100 is about 0.03 MB.

\medskip
\noindent\textbf{Comparison with FedAvg.}
With the above calculation as a base, we compare the per-round upstream communication cost incurred by FedAF and that of typical \textit{aggregate-then-adapt} method such as FedAvg in Table~\ref{tab:suppl-comm-cost}, where FedAF using an image-per-class (IPC) of 50. As described earlier, we use three random seeds to generate the three sets of data distribution and report the average communication overhead.
Note that the communication cost of transmitting $\mathcal{V}_k$, $\mathcal{V}$, and $\mathcal{R}_k$ is negligible compared to transmitting the condensed data and the model, so the downstream communication cost is essentially the same as that incurred by downloading the global model from the server, which is the same for FedAvg and FedAF. From Table~\ref{tab:suppl-comm-cost}, one can observe that for training the ResNet18 model, FedAF is much more efficient in communication cost compared to FedAvg. When learning the ConvNet model, which is relative smaller in size, FedAF still achieves significantly higher communication efficiency than FedAvg, especially on FMNIST and CIFAR10. While FedAvg incurs slightly less communication than FedAF for learning the ConvNet model on CIFAR100, FedAF drastically outperforms FedAvg in accuracy and convergence (see performance comparison in Section~\ref{ssec:results-label-noniid}). Moreover, unlike FedAvg, where the communication cost is solely determined by the model size and thus becomes increasingly expensive when a larger model is being learned, FedAF's communication cost is irrespective of the size of the underlying model. More interestingly, FedAF incurs less communication overhead in stronger label-skew data heterogeneity scenarios. These merits mark the extraordinary cost-effectiveness of FedAF.

\begin{table}[!t]
    \centering
    \setlength{\tabcolsep}{3.5pt}
    \fontsize{9pt}{9pt}\selectfont
    \begin{tabular}{c|c|c|c|c|c}
        \toprule
        \multirow{2}{*}{Dataset} & \multirow{2}{*}{$\alpha$}  & \multicolumn{2}{c}{CNN}                & \multicolumn{2}{c}{ResNet18} \\
                                 &                            & FedAvg                    & FedAF      & FedAvg     & FedAF   \\
        \midrule
        \multirow{3}{*}{FMNIST}  & 0.02                       & \multirow{3}{*}{1.46 MB}  & 0.06 MB    & \multirow{3}{*}{42.65 MB} & 0.06 MB\\
                                 & 0.05                       &                           & 0.09 MB    &                           & 0.09 MB\\
                                 & 0.1                        &                           & 0.14 MB    &                           & 0.14 MB\\
        \midrule
        \multirow{3}{*}{CIFAR10} & 0.02                       & \multirow{3}{*}{1.46 MB}  & 0.22 MB    & \multirow{3}{*}{42.65 MB} & 0.22 MB\\
                                 & 0.05                       &                           & 0.31 MB    &                           & 0.31 MB\\
                                 & 0.1                        &                           & 0.44 MB    &                           & 0.44 MB\\
        \midrule
        \multirow{3}{*}{CIFAR100}& 0.02                       & \multirow{3}{*}{1.46 MB}  & 1.93 MB    & \multirow{3}{*}{42.65 MB} & 1.93 MB\\
                                 & 0.05                       &                           & 2.46 MB    &                           & 2.46 MB\\
                                 & 0.1                        &                           & 3.22 MB    &                           & 3.22 MB\\
        \bottomrule    
    \end{tabular}
    \vskip -5 pt
    \caption{Per-round upstream communication cost incurred by FedAvg and FedAF for learning CNN and ResNet18 on FMNIST, CIFAR10, and CIFAF100. FedAF uses an IPC of 50.}
    \vskip -10 pt
    \label{tab:suppl-comm-cost}
\end{table}

\end{document}